%% file: main.tex
\documentclass[lettersize,journal]{IEEEtran}
\usepackage{amsmath,amsfonts}
\usepackage{algorithm}
\usepackage{array}
\usepackage[caption=false,font=normalsize,labelfont=sf,textfont=sf]{subfig}
\usepackage{textcomp}
\usepackage{stfloats}
\usepackage{url}
\usepackage{verbatim}
\usepackage{graphicx}
\usepackage{cite}
\usepackage{multirow}
\usepackage{times}
\usepackage{epsfig}

\usepackage{graphicx}
\usepackage{amsmath}
\usepackage{amssymb}
\usepackage{booktabs}
\usepackage{algorithm}
\usepackage{algpseudocode}
\usepackage{mathtools}
\usepackage{multirow}

\usepackage[capitalize]{cleveref}
\crefname{section}{Sec.}{Secs.}
\Crefname{section}{Section}{Sections}
\Crefname{table}{Table}{Tables}
\crefname{table}{Tab.}{Tabs.}

\begin{document}

\title{InfoGCN++: Learning Representation \\ by Predicting the Future for Online Human Skeleton-based Action Recognition}

\author{Seunggeun Chi$^{*1}$, Hyung-gun Chi$^{*1}$, Qixing Huang$^{2}$, Karthik Ramani$^{1}$
\thanks{$^{1}$Seunggeun Chi, Hyung-gun Chi, and Karthik Ramani are with the School of Electrical Engineering at Purdue University, West Lafayette, IN, USA. E-mail:  {\tt\small \{hgchi, sgchi, ramani\}@purdue.edu}}
\thanks{$^{2}$Qixing Huang is with the Department of Computer Science, University of Texas Austin, Austin, TX, USA. E-mail: {\tt\small huangqx@cs.utexas.edu}}.
\thanks{$^{*}$These authors contributed equally to this work.}
\thanks{This work has been submitted to the IEEE for possible publication. Copyright may be transferred without notice, after which this version may no longer be accessible.}}



\maketitle

\input{sections/0_abstract} \label{sec:abstarct}
\input{sections/1_introduction} \label{sec:introduction}
\input{sections/2_related_works} \label{sec:related_work} 
\input{sections/3_method} \label{sec:method} 
\input{sections/4_experiments} \label{sec:experiment}
\input{sections/5_conclusion} \label{sec:conclusion}

\newpage
\bibliography{egbib.bib, IEEEfull.bib}
\bibliographystyle{IEEEtrans}

\newpage

\vspace{11pt}

\vspace{11pt}

\vfill

\end{document}

%% file: sections/0_abstract.tex
\begin{abstract}
Skeleton-based action recognition has made significant advancements recently, with models like InfoGCN showcasing remarkable accuracy. However, these models exhibit a key limitation: they necessitate complete action observation prior to classification, which constrains their applicability in real-time situations such as surveillance and robotic systems. To overcome this barrier, we introduce InfoGCN++, an innovative extension of InfoGCN, explicitly developed for online skeleton-based action recognition. InfoGCN++ augments the abilities of the original InfoGCN model by allowing real-time categorization of action types, independent of the observation sequence's length. It transcends conventional approaches by learning from current and anticipated future movements, thereby creating a more thorough representation of the entire sequence. Our approach to prediction is managed as an extrapolation issue, grounded on observed actions. To enable this, InfoGCN++ incorporates Neural Ordinary Differential Equations, a concept that lets it effectively model the continuous evolution of hidden states. Following rigorous evaluations on three skeleton-based action recognition benchmarks, InfoGCN++ demonstrates exceptional performance in online action recognition. It consistently equals or exceeds existing techniques, highlighting its significant potential to reshape the landscape of real-time action recognition applications. Consequently, this work represents a major leap forward from InfoGCN, pushing the limits of what's possible in online, skeleton-based action recognition.
The code for InfoGCN++ is publicly available at \url{https://github.com/stnoah1/infogcn2} for further exploration and validation.
\end{abstract}

\begin{IEEEkeywords}
Human Action Recognition, Online Action Recognition, Human Motion Prediction, Neural ODE.
\end{IEEEkeywords}

%% file: sections/1_introduction.tex
\section{Introduction}

Human action recognition, a critical branch of computer vision, is indispensable in a wide array of applications, including but not limited to Augmented Reality (AR), Virtual Reality (VR) \cite{wang2021gesturar, huang2021adaptutar}, human-robot interaction \cite{pei2011parsing, vondrick2016anticipating}, and autonomous vehicles \cite{brehar2021pedestrian, pop2019multi}. Given the nature of these applications—whether designed to assist humans or perceive them as essential elements in a dynamic environment—it is important to have a robust and accurate understanding of human actions.

In this context, skeleton-based action recognition has proven to be particularly effective. The recent surge in its popularity can be attributed to its resilience against the cluttered background often encountered in video data and the advancements in 3D camera technologies. Many pioneering works have leveraged this approach, reporting impressive performance in classifying a variety of actions with a high degree of accuracy \cite{liu2020disentangling,chen2021channel,yan2018spatial,zhang2020semantics,shi2019two,shi2019skeleton,li2019actional,cheng2020skeleton,korban2020ddgcn,ye2020dynamic,chen2021multi}.

However, a significant limitation of existing methods, including the state-of-the-art method InfoGCN \cite{chi2022infogcn}, is the requirement to capture the entire action observation for classification, leading to latency issues in real-time applications. The need to capture the entire action observation as input for classification often results in a latency that can be up to 10 seconds. For instance, recognizing the action ``wear a shoe" from the NTU dataset can be time-consuming \cite{shahroudy2016ntu}. This latency impedes applications such as human-robot interaction, where swift anticipatory behavior is crucial for seamless cooperation \cite{van2020ergonomic, vondrick2016anticipating}.
 \IEEEpubidadjcol

\begin{figure}[t]
\centering
\includegraphics[width=1.0\linewidth]{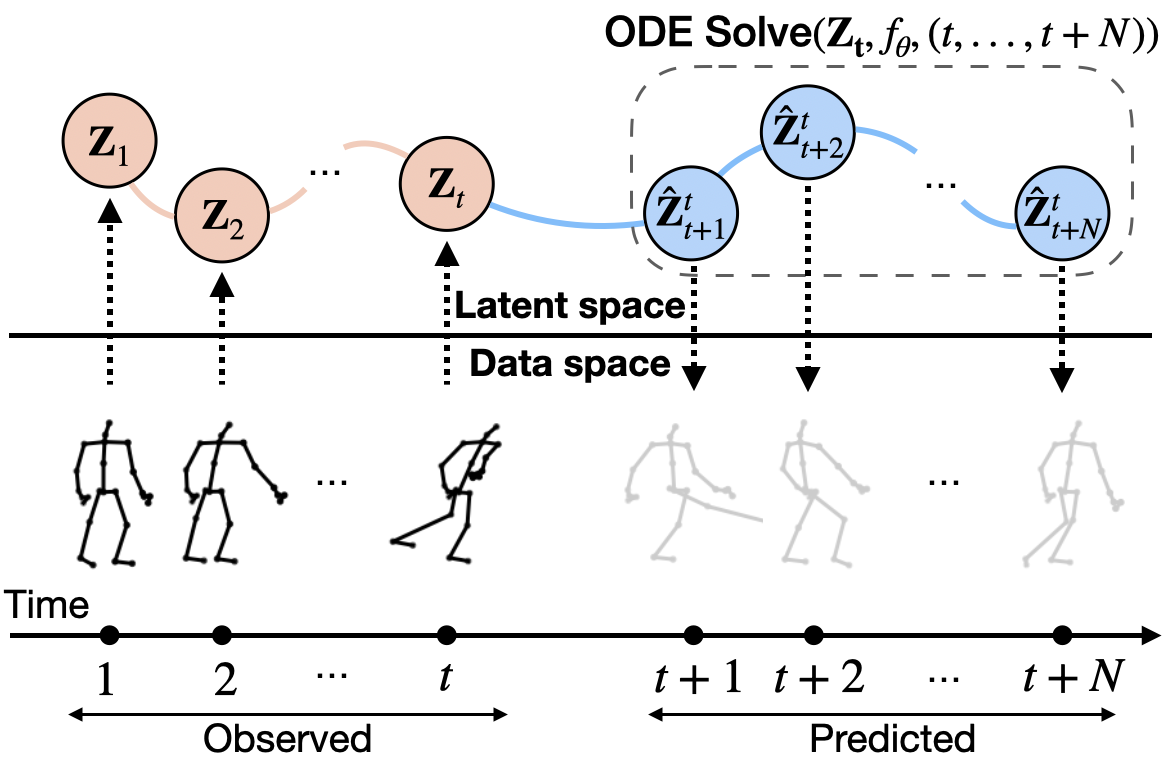}
\caption{
\textbf{A visual representation of the InfoGCN++ model.} The InfoGCN++ model leverages Neural ODE to predict future movements from given observations, thereby forming comprehensive sequence representations. This anticipatory approach equips the model with the necessary discriminative information for swift and accurate action recognition.}
\label{fig:intro}
\end{figure}

To address this challenge, we present InfoGCN++, an innovative extension to the original InfoGCN \cite{chi2022infogcn} that specifically targets online action recognition. By facilitating recognition as actions progressively unfold, InfoGCN++ eliminates the need for complete action observation prior to classification, thus paving the way for real-time applications. While InfoGCN++ inherits the successful skeleton embedding and spatial encoding components of InfoGCN, it departs from its predecessor by introducing a unique capability to recognize actions from partial observations. This capability addresses a key limitation of existing models \cite{foo2022era, wang2021dear, li2020hard, wang2019progressive, hu2018early, weng2020early} on early action recognition, which tends to predict actions at specific fractions of the observation (such as 10\% or 50\%), rather than during the continuous performance of the action. 

Our multi-task learning strategy, employed in InfoGCN++, simultaneously performs future motion prediction and action recognition. By learning to predict future motion, the model can better anticipate the future trajectory of the skeleton, thereby enhancing its ability to recognize actions in real-time. This is achieved by incorporating the concept of NeuralODE~\cite{chen2018neural}, which models the time series data dynamics by solving ordinary differential equations (ODEs) using a neural network, as shown in \cref{fig:intro}. By incorporating this advanced modeling approach, InfoGCN++ bridges the gap between observing actions and predicting their outcomes, significantly reducing the latency of the recognition process.

To rigorously evaluate InfoGCN++'s performance, we leverage several well-established datasets for skeleton-based action recognition, including NTU RGB+D 60~\cite{shahroudy2016ntu}, 120~\cite{liu2019ntu}, and NW-UCLA~\cite{wang2014cross}. These datasets provide a diverse range of action sequences, making them ideal for testing our approach.

Furthermore, to contextualize InfoGCN++'s performance, we draw comparisons with other prevalent methods in the field. Specifically, we compare ours with early action prediction methods and offline skeleton-based action recognition methods. This comparative evaluation is conducted using varying fractions of observation, offering a fair and comprehensive assessment of the strengths and weaknesses of each method.

Our results indicate that InfoGCN++ either surpasses or is on par with the existing methods in terms of performance. More importantly, InfoGCN++ provides a distinct advantage, it allows for continuous and real-time action inference. This capability not only confirms the effectiveness of our approach but also demonstrates its potential for transforming real-time applications in various fields.

Our contributions are summarized as follows:
\begin{itemize}
\item We propose a novel framework named InfoGCN++ for online skeleton-based action recognition that addresses the challenge of requiring the entire action sequence for classification in real-time applications.
\item InfoGCN++ utilizes multi-task learning to simultaneously learn action recognition and future motion prediction, improving its discriminative power for action recognition by anticipating the future movement of the skeleton.
\item InfoGCN++ achieves competitive performance compared to existing methods while having the advantage of continuous online action inference, making it well-suited for real-time applications where anticipatory behavior is crucial.
\end{itemize}

%% file: sections/2_related_works.tex
\section{Related Works}
\subsection{Offline Skeleton-based Action Recognition}

\noindent There has been a surge of interest in skeleton-based action recognition recently, resulting in significant strides in this domain~\cite{liu2020disentangling,chen2021channel,yan2018spatial,zhang2020semantics,shi2019two,shi2019skeleton,li2019actional,cheng2020skeleton,korban2020ddgcn,ye2020dynamic,chen2021multi}. Many approaches seek to enhance performance by leveraging Graph Convolution Networks to model the topology between the joints~\cite{kipf2016semi}. A subset of this work~\cite{li2018spatio, yan2018spatial, shi2019skeleton, liu2020disentangling} focuses on exploiting the explicit topology of the joints, using the skeleton as a graph to extract features. Meanwhile, other studies~\cite{li2019actional, shi2019two, chen2021channel, chi2022infogcn} delve into the modeling of intrinsic topology to further uncover implicit connections between joints. Despite the advances made, these methods are unsuitable for applications requiring low latency. A method for online skeleton-based action recognition is proposed by Liu \textit{et al.} \cite{liu2018online}; however, it primarily concentrates on detecting short action clips from untrimmed videos. In contrast, our work tackles the challenge of continuously recognizing ongoing actions from trimmed videos that can span over 10 seconds.

\begin{table}[t]
\centering
\caption{Summary of different action recognition tasks.}
\resizebox{1.0\linewidth}{!}{%
\begin{tabular}{|l|l|l|}
\hline
\multicolumn{1}{|c}{Task}     & \multicolumn{1}{|c|}{Observation} & \multicolumn{1}{c|}{Inference Frequency} \\
\hline\hline
Online Action Recognition    & Partial & Every frame                    \\ \hline
Early Action Prediction      & Partial & Specific fractions           \\ \hline
Offline Action Recognition & Whole & One time                    \\ \hline
\end{tabular}
}
\label{table:task}
\end{table}

\subsection{Skeleton-based Early Action Prediction}
\noindent Early action prediction~\cite{weng2020early, foo2022era, li2020hard, xu2019temporal, wang2021dear, stergiou2022temporal, hu2018early, wang2019progressive} is another active area of research that aims to predict actions as early as possible based on the initial parts of a video. Notable works in this direction~\cite{foo2022era, wang2021dear, li2020hard, wang2019progressive, hu2018early, weng2020early} primarily focus on skeleton data. Several innovative methods have been proposed; for example, Foo \textit{et al.} \cite{foo2022era} and Liu \textit{et al.} \cite{li2020hard} focus on distinguishing subtle differences between challenging samples, while Wang \textit{et al.} \cite{wang2019progressive} employ a teacher-student distillation model to transfer long-term knowledge. However, as these models are designed to recognize actions based on a specific fraction of the observation, they are not suitable for recognizing streaming actions online. In Table \ref{table:task}, we present a comparison of the action recognition tasks to emphasize the distinctions between the existing tasks and our online action recognition task.
In contrast to early action prediction works, our proposed InfoGCN++ is designed to recognize ongoing actions where observations evolve continuously. We achieve this through the novel use of NeuralODEs, enabling us to predict future motion for each evolving observation.

\subsection{Neural Ordinary Differential Equation}
\noindent Chen \textit{et al.} \cite{chen2018neural} first introduced the concept of Neural Ordinary Differential Equation (NeuralODE), a mechanism whose output is a black-box differential equation solver. This method offers a generative approach to modeling time-series data, representing them via continuous latent trajectories. The potential of NeuralODEs has been explored further in the context of irregularly sampled time-series data by Rubanova \textit{et al.} \cite{rubanova2019latent}, while Park \textit{et al.} \cite{park2021vid} have investigated video generation using the same concept. Chi \textit{et al.} \cite{chi2023adamsformer} delved into its applications to localize action in the future frames. In our study, we leverage the concept of NeuralODE to predict future skeleton motion, which facilitates the recognition of actions from partial observations. This unique application of NeuralODEs sets our work apart from existing research and presents a novel direction in the domain of online action recognition.

%% file: sections/3_method.tex
\begin{figure*}[t]
    \centering
    \includegraphics[width=0.95\linewidth]{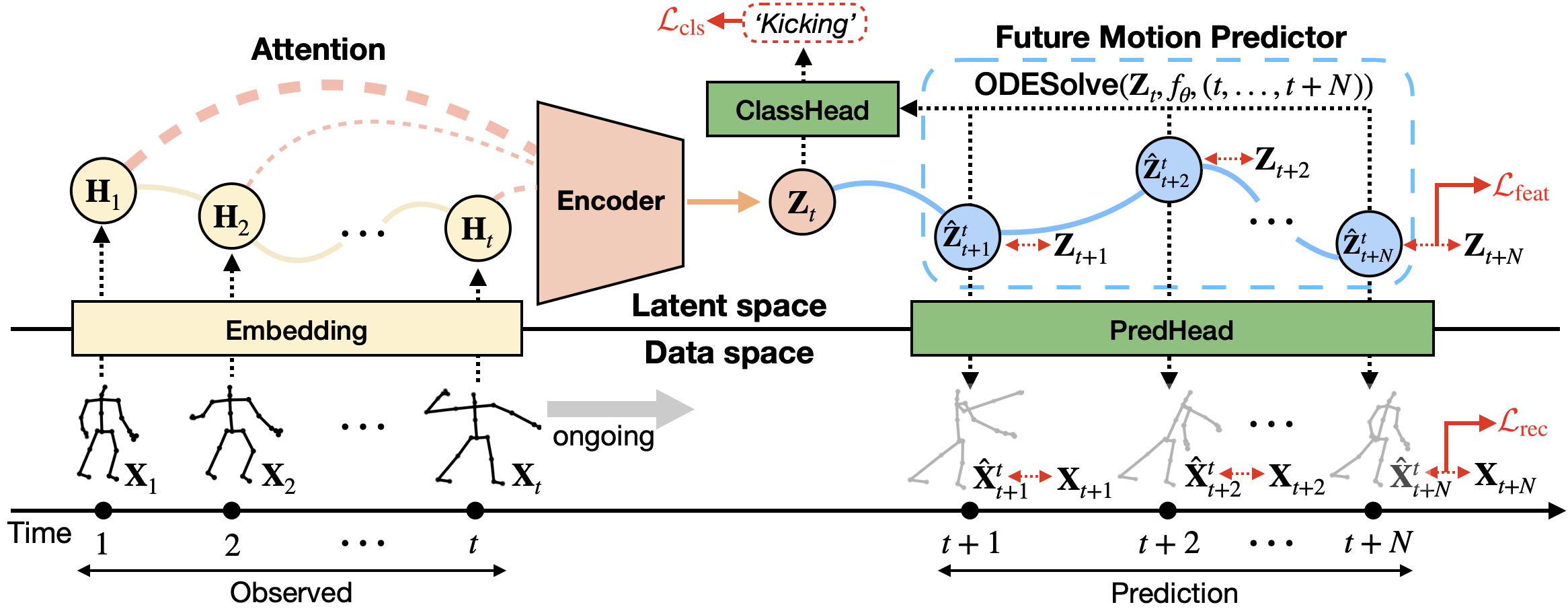}
    \caption{\textbf{Overview of proposed InfoGCN++.} Given the representation of the observation $\mathbf{Z}_t$, the InfoGCN++ extrapolates the representation to future frames by solving the IVP to predict future motion. Learned representations by predicting the future are then used for classifying the action at a given observation. The detailed structures for the encoder and classification head are shown in \cref{fig:architecture}.}
    \label{fig:framework}
\end{figure*}

\section{Online Skeleton-based Action Recognition}\label{Section:Problem:Statement}
\noindent Online skeleton-based action recognition is a real-time task that involves classifying actions from a streaming skeleton action sequence. The goal is to predict the action category $y$ from every accumulated partial observation $\mathbf{X}_{1:t}$ up to time $t$, where $\mathbf{X}_t \in \mathbb{R}^{V\times 3}$ is the 3D skeleton pose at time $t$, and $V$ is the number of joints in the skeleton. Unlike early action prediction, which predicts actions based on a limited fraction of observations (i.e., 10\%, 50\%), online action recognition requires inference of the action in every single frame with low latency, making it more efficient and adaptable for real-time applications. We summarize the difference between the tasks in \Cref{table:task}.

\section{Preliminaries}
\subsection{Initial Value Problem and NeuralODE}
\noindent A NeuralODE \cite{chen2018neural} is an ordinary differential equation (ODE) where a neural network formulates its derivative function.
NeuralODE is used to solve the initial value problem (IVP), which is defined as ODE with an initial value $h_1$:
\begin{flalign}
    \frac{dh(t)}{dt}=f_\theta(h(t),t), \text{ where } h(t_1) = h_1,
\end{flalign}
where the neural network $f_\theta(\cdot)$ models the first-order derivative of $h(t)$ with a learnable parameter $\theta$.
By using numerical methods such as the Euler method or Runge-Kutta, a hidden state $h_t$ for future time $t$ can be derived by recursively predicting the next hidden state:
\begin{flalign}
    h_1, \ldots, h_{\tau} = \text{ODESolve}(f_\theta, h_1, (t_1,  \ldots, t_{\tau})).\label{eq:ode} 
\end{flalign}

\subsection{Self-Attention Graph Convolution.}
\noindent The Graph Convolution Network (GCN) is a powerful model that enables efficient extraction of features from graph-structured data~\cite{kipf2016semi}. Considering the human skeleton, which can be represented as a graph where the nodes symbolize joints and edges signify bones, the GCN serves as a robust mechanism for capturing the spatial characteristics of the skeletal structure.
In this context, we can denote a binary adjacency matrix as $\mathbf{A} \in \mathbb{R}^{V \times V}$, and a feature matrix of the graph with $V$ nodes and $D$-dimensional node features as $\mathbf{H} \in \mathbb{R}^{V \times D}$. The operation of the GCN is defined as,
\begin{flalign}
\text{GCN}(\mathbf{H}) = \sigma(\sum_{m=1}^{M}\tilde{\mathbf{A}}\mathbf{H}\mathbf{W}_m), \label{eq:gcn}
\end{flalign}
Here, $\sigma(\cdot)$ represents a non-linear activation function such as ReLU~\cite{agarap2018deep}, $\mathbf{H}$ is the feature matrix, $M$ symbolizes the number of heads, and $\mathbf{W} \in \mathbb{R}^{D\times D}$ is a learnable weight matrix. In the base version of GCN~\cite{kipf2016semi}, a symmetrically normalized form of $\mathbf{A}$ is utilized for $\tilde{\mathbf{A}}$.

\begin{figure*}[t]
    \centering
    \includegraphics[width=0.95\linewidth]{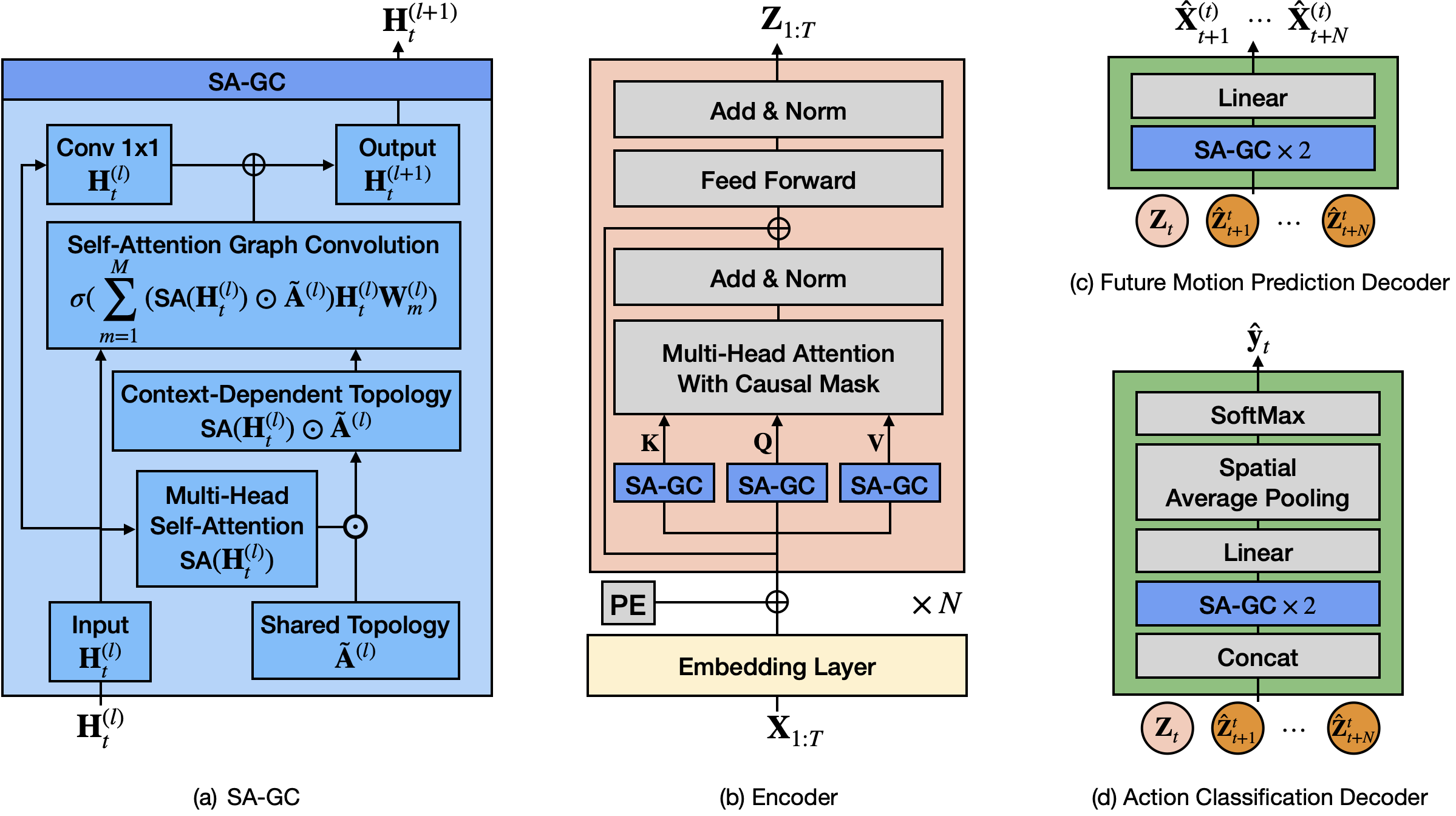}
    \caption{The detail architecture of the \textbf{(a)} SA-GC \cite{chi2022infogcn} module, \textbf{(b)} the Encoder, \textbf{(c)} Future motion prediction decoder, and \textbf{(d)} Action classification decoder of infoGCN++.}
    \label{fig:architecture}
\end{figure*}

InfoGCN \cite{chi2022infogcn} introduced the concept of Self-Attention Graph Convolution (SA-GC), which incorporates a self-attention-based adjacency matrix into graph convolution. SA-GC has proven its effectiveness in encoding human action by discerning the context-dependent intrinsic topology of the human skeleton. In our study, we deploy SA-GC as a spatial encoding method for the human skeleton.
The neighborhood information for graph convolution is represented by SA-GC as $\tilde{\textbf{\text{A}}}_m \odot \text{SA}_m(\textbf{\text{H}}_t)$ where $\odot$ indicates Hadamard product. The overall rule for updating the joint representation is expressed as,
\begin{flalign}
\text{SA-GC}(\textbf{H}) = \sigma\bigg(\sum_{m=1}^{M}\big(\tilde{\textbf{\text{A}}}_m \odot \text{SA}_m(\textbf{\text{H}})\big) \textbf{\text{H}}\textbf{\text{W}}_{m}\bigg),\label{eq:sagc} 
\end{flalign}
where $\text{SA}(\textbf{\text{H}}) = \texttt{softmax}(\textbf{\text{H}} \textbf{\text{W}}_K (\textbf{\text{H}} \textbf{\text{W}}_Q)^\top/\sqrt{D'})$ and $\textbf{\text{W}}_Q, \textbf{\text{W}}_K \in \mathbb{R}^{D\times D'}$ are the linear projections of $\mathbf{H}$ to Key and Queue of a self-attention. A multi-head attention approach with the number of heads $M$ is employed, following InfoGCN. The overall SA-GC process is illustrated in \ref{fig:architecture} (a)
\section{InfoGCN++}
\noindent In this section, we introduce InfoGCN++, a carefully engineered extension of InfoGCN  \cite{chi2022infogcn} tailored to tackle online skeleton-based action recognition. The cornerstone of InfoGCN++ is its ability to construct a richly informed representation for each observed action, thus providing an accurate foundation for subsequent action classification. In addition to this core function, InfoGCN++ embarks on an auxiliary task of future motion prediction. This innovative feature equips InfoGCN++ with the capacity to extrapolate a complete sequence representation even from partial observations. This twofold approach not only enhances the model's action recognition capability but also situates it as a future-ready solution for real-time applications.

\subsection{Architecture overview}
\noindent InfoGCN++ comprises four key components: an embedding layer, an encoder, a future motion predictor, and task-specific decoders.
The embedding layer, encoder, and action classification decoder are inherited from InfoGCN, while the future motion predictor and future motion prediction decoder is a newly introduced component exploiting SA-GC of InfoGCN \cite{chi2022infogcn}.
The embedding layer transforms the 3D skeleton data into latent space, and the encoder creates a sequence of representations encapsulating the spatiotemporal features of the skeleton. Leveraging an initial value problem (IVP) formulation, the future motion predictor extrapolates future motion based on the current observation. The extrapolated representation is used for both future motion prediction and action category classification tasks.

\subsection{Embedding Layer}
\noindent Following the InfoGCN, the embedding layer transforms the 3D skeleton feature $\mathbf{X}_t \in \mathbb{R}^{V \times 3}$ into a $D$ dimensional latent feature through linear projection.
We then add learnable spatial positional embeddings $\text{PE} \in \mathbb{R}^{V\times D}$ to inject the positional information of joints.
The overall skeleton embedding process is formulated as,
\begin{flalign}
    \textbf{\text{H}}_t^{(0)} = \texttt{Linear}(\textbf{\text{X}}_t) + \text{PE}, \label{eq:joint_embedding}
\end{flalign}

\subsection{Encoder}
\noindent The encoder is designed to convert the latent features of the skeleton pose from time 1 to $t$ ($\mathbf{H}_{1:t}$) to representation $\mathbf{Z}_t \in \mathbb{R}^{V \times D}$:
\begin{flalign}
\mathbf{Z}_{t} = \text{Encoder}(\mathbf{H}_{1:t}).
\end{flalign}
Our encoder is comprised of two fundamental components: a spatial modeling module and a temporal modeling module. Although the overarching encoding procedure takes inspiration from InfoGCN \cite{chi2022infogcn}, we deviate from its approach by employing the Transformer model \cite{vaswani2017attention} for the purpose of temporal encoding. With a casual mask for temporal modeling, we strategically guide the model to focus only on historical frames, further optimizing our system for real-time, online skeleton-based action recognition.

We adopt the Transformer \cite{vaswani2017attention} encoder with a causal mask for temporal modeling. The causal mask enables our model to attend only observed frames, making it capable of online action recognition.
To combine spatial and temporal features, we replace the linear projection for Queue $\mathbf{Q}_t$, Key $\mathbf{K}_t$, and Value $\mathbf{V}_t$ embedding in Transformer with SA-GC (\ref{eq:sagc}) of infoGCN \cite{chi2022infogcn} to extracts contextual features from the skeleton action (See \ref{fig:architecture} (b)).

We calculate temporal attention between hidden feature $\mathbf{H}^{(l)}_t$ and those of all previous observations $\mathbf{H}^{(l)}_{1:t}$. 
We only calculate attention between the different times at the same joint index $i$ because we aggregate spatial features from the projection layer using GCN.
Then, the attention process of our encoding layer is defined as follows,
\begin{flalign}
    \text{Attention}_{t}[i] = \text{Softmax(}\frac{\mathbf{Q}_{t}[i]\mathbf{K}_{1:t}^\top[i]}{\sqrt{D}})\mathbf{V}_{1:t}[i],
\end{flalign}
where $1\leq i \leq V$.
Further, we use multi-head self-attention (MHSA) followed by multi-layer perceptron (MLP), and we employ layer normalization \cite{ba2016layer} (Layer Norm) for inputs for both MHSA and MLP.
The detailed architecture of the encoder is illustrated in \cref{fig:architecture} (a).

\subsection{Future Motion Predictor}
\noindent The future motion predictor predicts the representation of future $N$ frames ($\mathbf{\hat{Z}}^{t}_{t+1:t+N}$) given the representation of observation $\mathbf{Z}_{t}$ from the encoder.
To predict future motion representation, we reformulate the prediction problem as an extrapolation problem.
Motivated by NeuralODE \cite{chen2018neural}, which successfully models the dynamics of continuous latent flow, we define the Initial Value Problem (IVP) with the representation of an observation $\mathbf{Z}_t$ as an initial value to extrapolate representations to the future $N$ frames.
With the ODE function $f_{\theta}(\cdot)$ that captures the dynamics of latent features, we solve the IVP to derive extrapolated future representations using the ODE solver:
\begin{flalign}
    \mathbf{\hat{Z}}_{t:t+N}^{(t)} = \text{ODESolve}(f_\theta, \mathbf{Z}_{t}, (t, ..., t+N)) \label{eq:odesolve}
\end{flalign}
Here, we denote $\mathbf{\hat{Z}}_{t+n}^t$ as the predicted representation of the $n$-th future frame from ODESolve with an initial condition $\mathbf{Z}_t$, and $\mathbf{\hat{Z}}_{t}^t = \mathbf{Z}_t$.
We use the Runge-Kutta method for the ODE solver.
We tried other ODE solvers like Mid-point and Dormand-Prince, but they have little difference in terms of motion prediction loss.

We design the ODE function $f_\theta(\cdot)$ with a neural network following \cite{chen2018neural}. For the time-variant property, we first add temporal positional embedding ($\text{PE}_{\text{temporal}} \in \mathbb{R}^{T \times D}$) to the input of the ODE function.
Specifically, we use sinusoidal positional embedding \cite{vaswani2017attention} for $\text{PE}_{\text{temporal}}$ to inject relative temporal positional information.
Then, the feature is passed through multiple SA-GC layers to model the dynamics of latent states. We stack two SA-GC layers to construct the ODE function. 

\subsection{Task-specific Decoders}\label{sec:head}
\noindent InfoGCN++ uses multi-task learning to simultaneously learn action recognition and future motion prediction. The future motion prediction task guides the model to anticipate future movements, enhancing its discriminative power for action recognition. By learning both tasks simultaneously, InfoGCN++ can represent the entire action sequence from partial observation and achieve better accuracy and efficiency in action recognition.

\subsubsection{Future Motion Prediction Decoder}
\begin{figure}[t]
  \centering
  \includegraphics[width=1.0\linewidth]{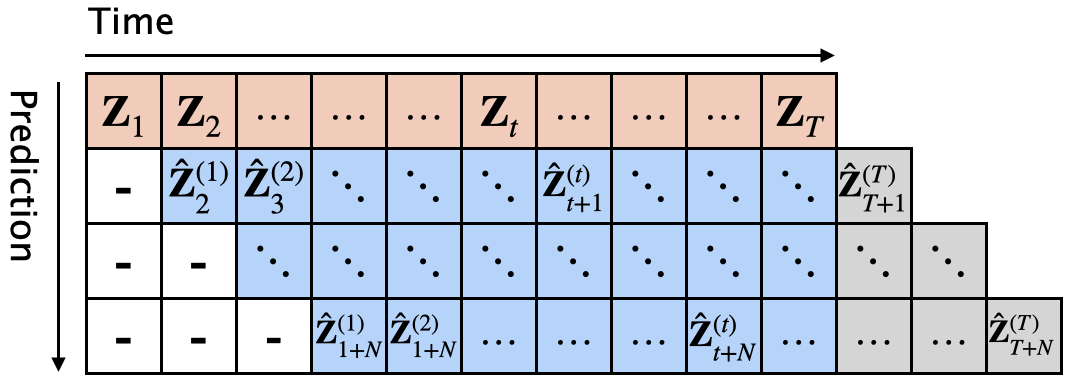}
  \caption{The visualization of a representation $\mathbf{Z}_{t}$ (\textit{first row}) and predicted future representations $\mathbf{\hat{Z}}^{(t)}_{t+1:t+N}$.}
  \label{fig:NXT}
\end{figure}

\noindent We propose a future motion prediction task as an auxiliary task for action classification in InfoGCN++. This task guides the model to capture the dynamics of the skeletons by predicting the trajectory of $N$ future frames. It helps the encoder to encode a better representation for action classification, representing the entire sequence from partial observation. The future motion prediction head comprises two GCN layers followed by a Linear layer that projects the latent feature to 3D data space (see \cref{fig:architecture} (a)). By using this approach, InfoGCN++ can learn to represent the entire action sequence from partial observation and anticipate future motions, enhancing its discriminative power for action recognition.
\begin{flalign}
\hat{\mathbf{X}}^{(t)}_{t+n} = \text{PredHead}(\hat{\mathbf{Z}}^{(t)}_{t+n}).
\end{flalign}
We define the motion prediction loss using the Mean Square Error (MSE) between the ground truth and prediction.
\begin{flalign}
\hspace{-.8em}\mathcal{L}_{\text{pred}} = \frac{1}{K}\sum_{n=1}^N\sum_{t=1}^{T-n}{\text{MSE}(\hat{\mathbf{X}}^{(t)}_{t+n}, \mathbf{X}_{t+n})},\!
\end{flalign}
where $K\!\!=\!\!NT\!-\! N(N\!+\!1)/2$.
Since the predicted representations where $t+n > T$ have no ground truth (gray boxes in \cref{fig:NXT}), we omit those frames when calculating the loss.

Further, we supervise the model to predict future $\mathbf{\hat{Z}}^{(t)}_{t+n}$ matching with the true future representation $\mathbf{Z}_{t+n}$ that comes from encoding the future observations ($\mathbf{Z}_{t+n}=\text{Encoder}(\mathbf{H}_{1:t+n})$). 
The design of feature loss is motivated by previous works \cite{girdhar2021anticipative, han2019video, han2020memory} that show anticipating future representations is practical self-supervision.
We use MSE to define the feature loss.
\begin{flalign}
\mathcal{L}_{\text{feat}} = \frac{1}{K}\sum_{n=1}^N\sum_{t=1}^{T-n}{\text{MSE}(\mathbf{\hat{Z}}^{(t)}_{t+n},\mathbf{Z}_{t+n})}.
\end{flalign}

\subsubsection{Action Classification Decoder}
\noindent Through the classification task, the predicted representation embeds the information to classify the action.
The classification head consists of two SA-GC layers followed by a Linear layer, spatial mean pooling, and SoftMax function that converts logits to categorical distribution as illustrated in \cref{fig:architecture} (b).
We concatenate the predicted representations for the classification head input.
\begin{flalign}
    \hat{\mathbf{y}}_{t} = \text{ClassHead}(\hat{\mathbf{Z}}^{(t)}_{t:t+N}).
\end{flalign}
Classification loss is defined using cross-entropy:
\begin{flalign}
    \mathcal{L}_{\text{cls}} = - \frac{1}{TC}\sum_{t=1}^T\sum_{c=1}^C{\mathbf{y}[c]\log([\hat{\mathbf{y}}_t[c])},
\end{flalign}
where $C$ is the number of action categories, and $\mathbf{y}$ is a one-hot encoded vector of ground truth. We further employ label smoothing \cite{szegedy2016rethinking} with a value of 0.1 for generalization.

\subsection{Training}
\noindent We combine the prediction, feature, and classification losses to train our model. The total loss is defined as,
\begin{flalign}
\mathcal{L} = \mathcal{L}_{\text{cls}}  + \lambda_1 \mathcal{L}_{\text{pred}} + \lambda_2 \mathcal{L}_{\text{feat}},
\end{flalign}
where $\lambda_1$ and $\lambda_2$ are weight coefficients for feature loss and classification loss, respectively. 
We tested different combinations of coefficients and chose the set that showed the best performance. 
We compare our model performance on different coefficient combinations in the \Cref{table:coefficient}.

The overall training and inference procedure is summarized in \Cref{alg:training} and \Cref{alg:inference}, respectively.
We omit the embedding layer in the algorithms for simplicity.
To boost training on different observation lengths, we parallelize the training by exploiting the causal mask on the encoder, forcing the model to attend to only previous frames.
For inference, the motion prediction head is not used.

%% file: sections/4_experiments.tex
\section{Experiment}
\noindent To demonstrate the superiority of InfoGCN++ in online skeleton-based action recognition, we conducted experiments on three widely used datasets: NTU RGB+D 60 \cite{shahroudy2016ntu}, 120 \cite{liu2019ntu}, and NW-UCLA \cite{wang2014cross}. We compared InfoGCN++'s performance with 1) early action prediction methods, 2) offline skeleton-based action recognition methods, and 3) offline methods separately trained on different partial observations. 

\subsection{Datasets} 
\noindent For our experiments, we utilize three widely used skeleton action datasets.
\subsubsection{NTU RGB+D~\cite{shahroudy2016ntu, liu2019ntu}}
\noindent NTU RGB+D 60 \cite{shahroudy2016ntu} is a large-scale 3D human activity dataset comprising 56,880 videos across 60 distinct action classes. An extension to this dataset, named NTU RGB+D 120~\cite{liu2019ntu}, includes an additional 57,600 videos with 60 new action classes. We followed existing literature~\cite{chi2022infogcn, chen2021channel} to evaluate the real-time skeleton-based action recognition performance in different scenarios: cross-subject and cross-view for NTU RGB+D 60, and cross-subject and cross-setup for NTU RGB+D 120. For the sake of clarity, we denote cross-subject, cross-view, and cross-setup splits as X-Sub, X-View, and X-Set, respectively.

\setlength{\textfloatsep}{0em}
\begin{algorithm}[t]
\caption{Training Procedure}\label{alg:training}
\textbf{Input: }$\mathbf{X}_{1:T}\text{: Full observation of skeleton sequence.}$\\
\textbf{Output: }$\mathbf{\tilde{y}}_{1:T}\text{: Action category of each observation.}$

  \begin{algorithmic}[0]
        \For{$epoch \gets 1 \,\textbf{to} \,MaxEpoch$}
            \For{$t \gets 1 \,\textbf{to} \,T$} \Comment{Working parallel}
                \State $\mathbf{Z}_{t} \gets \text{Encoder}(\mathbf{X}_{1:t})$ 
                \State $\mathbf{\hat{Z}}_{t:t+N}^t \gets \text{ODESolve}(f_\theta, \mathbf{Z}_{t}, (t, ..., t+N))$
                \State $\hat{\mathbf{X}}^t_{t:t+N} \gets \text{PredDecoder}(\mathbf{\hat{Z}}_{t:t+N}^t)$
                \State $\hat{\mathbf{y}}_t \gets \text{ClassDecoder}(\mathbf{\hat{Z}}_{t:t+N}^t)$
            \EndFor
        \EndFor
    \end{algorithmic}
\end{algorithm}

\setlength{\textfloatsep}{1em}
\begin{algorithm}[t]
\caption{Inference Procedure}\label{alg:inference}
\textbf{Input: }$\mathbf{X}_{1:t}\text{: Accumulated observation of skeleton sequence.}$ \\
\textbf{Output: }$\mathbf{\tilde{y}}_{t}\text{: Action category.}$
    \begin{algorithmic}
        \State $\mathbf{Z}_{t} \gets \text{Encoder}(\mathbf{X}_{1:t})$
        \State $\mathbf{\hat{Z}}_{t:t+N}^t \gets \text{ODESolve}(f_\theta, \mathbf{Z}_{t}, (t, ..., t+N))$
        \State $\hat{\mathbf{y}}_t \gets \text{ClassDecoder}(\mathbf{\hat{Z}}_{t:t+N}^t)$
    \end{algorithmic}
\end{algorithm}

\begin{table*}[t]
\centering
\caption{Performance comparisons at different observation ratios. \\ \textit{(Left)} A single model trained using the full observation, and \textit{(Right)} multiple models trained using each observation.}
    \begin{tabular}{c l | r r r r r | r r r r r }
        \hline
        \multirow{3}{*}{Datasets} & \multirow{3}{*}{Methods} & \multicolumn{5}{c}{\textbf{Single model trained on 100\% OR}} & \multicolumn{5}{c}{\textbf{Separately trained on each OR}}\\
        & & \multicolumn{5}{c}{Observation Ratio (OR)} & \multicolumn{5}{c}{Observation Ratio (OR)}\\
        \cline{3-7} \cline{8-12}
         &  & 20\%  & 40\% & 60\%& 80\% & 100\% & 20\% & 40\% & 60\%& 80\% & 100\%\\
        \hline\hline
        \multirow{6}{*}{\begin{tabular}[c]{@{}c@{}}NTU 60 X-Sub\end{tabular}} & ST-GCN\cite{yan2018spatial} & 
                                                \enspace 10.78 & 33.44 & 67.68 & 78.34 & 81.50 
                                                & 37.36 & 64.74 & 76.32 & 80.99 & 81.50 \\
        & 2S-AGCN~\cite{shi2019two} &            \enspace 12.34 & 33.54 & 61.88 & 81.45 & 86.73  
                                                & 37.73 & 69.41 & 80.61 & 84.91 & 86.73 \\
        & MS-G3D~\cite{liu2020disentangling} &   \enspace 12.09 & 36.58 & 66.76 & 83.88 & 88.57  
                                                & 43.09 & 73.35 & 82.76 & 87.40 & 88.57 \\
        & CTR-GCN~\cite{chen2021channel} &       \enspace  9.02 & 33.93 & 66.32 & 86.07 & \textbf{89.93} 
                                                & 42.17 & 73.74 & 83.90 & \textbf{88.83} & \textbf{89.93} \\
        & InfoGCN~\cite{chi2022infogcn} &        \enspace 12.14 & 35.65 & 70.65 & \textbf{87.33} & 89.80 
                                                & 41.96 & \textbf{74.44} & \textbf{84.84} & 88.55 & 89.80 \\
        \cline{2-12}
        & \textbf{InfoGCN++ (Ours)} & \textbf{44.57} & \textbf{73.59} & \textbf{81.68} & 84.42 & 85.38 & \textbf{44.57} & 73.59 & 81.68 & 84.42 & 85.38 \\
        \hline
        \multirow{6}{*}{\begin{tabular}[c]{@{}c@{}}NTU 60 X-View\end{tabular}} & ST-GCN\cite{yan2018spatial} &
                                                \enspace 14.02 & 35.81 & 70.85 & 86.55 & 88.00 
                                                & 45.20 & 79.55 & 84.51 & 87.45 & 88.00 \\
        & 2S-AGCN~\cite{shi2019two} &            \enspace 14.25 & 38.34 & 69.57 & 89.57 & 93.95  
                                                & 55.15 & 84.08 & 89.96 & 92.93 & 93.95 \\
        & MS-G3D~\cite{liu2020disentangling} &   \enspace 11.34 & 38.35 & 73.33 & 90.65 & 95.00  
                                                & 53.46 & 86.76 & 90.83 & 93.87 & 95.00 \\
        & CTR-GCN~\cite{chen2021channel} &       \enspace 11.07 & 35.10 & 71.37 & 91.60 & 94.67  
                                                & 53.31 & 85.43 & 91.94 & 94.06 & 94.67 \\
        & InfoGCN~\cite{chi2022infogcn} &        \enspace 12.59 & 39.09 & 75.41 & \textbf{92.50} & \textbf{95.20} 
                                                & 54.56 & \textbf{86.88} & \textbf{93.09} & \textbf{94.42} & \textbf{95.20} \\
        \cline{2-12}
        & \textbf{InfoGCN++ (Ours)} & \textbf{55.57} & \textbf{84.62} & \textbf{90.79} & 92.35 & 92.55  & \textbf{55.57} & 84.62 & 90.79 & 92.35 & 92.55 \\
        \hline
        \multirow{6}{*}{\begin{tabular}[c]{@{}c@{}}NTU 120 X-Sub\end{tabular}} & ST-GCN\cite{yan2018spatial} & 
                                                \enspace  7.09 & 19.02 & 43.04 & 67.74 & 77.44  
                                                & 29.41 & 56.90 & 71.77 & 76.40 & 77.44 \\
        & 2S-AGCN~\cite{shi2019two} &            \enspace 10.59 & 28.67 & 55.50 & 74.34 & 80.36 
                                                & 33.10 & \textbf{61.51} & 73.74 & 76.70 & 80.36 \\
        & MS-G3D~\cite{liu2020disentangling} &   \enspace  7.94 & 24.25 & 52.46 & 76.08 & 83.41  
                                                & 31.00 & 61.45 & \textbf{78.23} & 82.13 & 83.41 \\
        & CTR-GCN~\cite{chen2021channel} &       \enspace  5.08 & 20.64 & 54.09 & \textbf{77.91} & 84.93 
                                                & 32.54 & 60.60 & 76.39 & 82.01 & 84.93 \\
        & InfoGCN~\cite{chi2022infogcn} &        \enspace  5.77 & 18.86 & 50.17 & 77.13 & \textbf{85.10} 
                                                & 31.72 & 60.70 & 76.42 & \textbf{82.54} & \textbf{85.10} \\
        \cline{2-12}
        & \textbf{InfoGCN++ (Ours)} & \textbf{33.63} & \textbf{58.86} & \textbf{71.72} & 76.35 & 77.42  & \textbf{33.63} & 58.86 & 71.72 & 76.35 & 77.42 \\
        \hline
        \multirow{6}{*}{\begin{tabular}[c]{@{}c@{}}NTU 120 X-Set\end{tabular}} & ST-GCN\cite{yan2018spatial} & 
                                                \enspace   8.35 & 21.37 & 46.20 & 71.74 & 80.36 
                                                & 29.67 & 60.32 & 72.77 & 77.45 & 80.36 \\
        & 2S-AGCN~\cite{shi2019two} &            \enspace 10.95 & 30.18 & 57.02 & 76.98 & 82.53 
                                                & 35.05 & 63.80 & 75.97 & 80.92 & 82.53 \\
        & MS-G3D~\cite{liu2020disentangling} &   \enspace  8.78 & 26.46 & 55.80 & 78.23 & 84.79 
                                                & 36.13 & \textbf{66.59} & \textbf{79.23} & 83.00 & 84.79 \\
        & CTR-GCN~\cite{chen2021channel} &       \enspace  9.02 & 21.02 & 51.99 & 78.63 & 85.93 
                                                & 32.54 & 60.60 & 78.67 & 84.40 & 85.93 \\
        & InfoGCN~\cite{chi2022infogcn} &        \enspace  7.59 & 25.67 & 58.47 & \textbf{80.58} & \textbf{86.30} 
                                                & 35.32 & 65.58 & 78.97 & \textbf{84.53} & \textbf{86.30} \\
        \cline{2-12}
        & \textbf{InfoGCN++ (Ours)} & \textbf{38.70} & \textbf{64.90} & \textbf{76.01} & 80.31 & 81.20  & \textbf{38.70} & 64.90 & 76.01 & 80.31 & 81.20 \\
        \hline
        \multirow{6}{*}{NW-UCLA} & ST-GCN\cite{yan2018spatial}&
                                                 \enspace 12.93 & 33.84 & 57.54 & 78.45 & 87.50 
                                                & 64.80 & 75.33 & 81.90 & 86.96 & 87.50 \\
        & 2S-AGCN~\cite{shi2019two} &            \enspace 24.57 & 51.94 & 68.53 & 83.19 & 90.87 
                                                & 62.28 & 76.29 & 86.64 & 87.28 & 90.87 \\
        & MS-G3D~\cite{liu2020disentangling} &   \enspace 26.94 & 53.45 & 75.65 & 88.79 & 92.67
                                                & \textbf{78.45} & 84.78 & 89.22 & 91.22 & 92.67 \\
        & CTR-GCN~\cite{chen2021channel} &       \enspace 13.79 & 43.32 & 71.55 & \textbf{91.16} & \textbf{95.04}
                                                & 76.94 & 83.84 & \textbf{90.73} & 91.59 & \textbf{95.04}\\
        & InfoGCN~\cite{chi2022infogcn} &        \enspace 23.71 & 57.54 & 79.74 & 90.52 & 94.00 
                                                &71.77 & 82.11 & 86.64 & \textbf{91.81} & 94.00 \\
        \cline{2-12}
        & \textbf{InfoGCN++ (Ours)} & \textbf{78.02} & \textbf{85.99} & \textbf{88.15} & 90.73 & 90.09 & 78.02 & \textbf{85.99} & 88.15 & 90.73 & 90.09 \\
        \hline
    \end{tabular}    
\label{table:sota}
\end{table*}

\begin{figure}[t]
    \centering
    \includegraphics[width=1.0\linewidth]{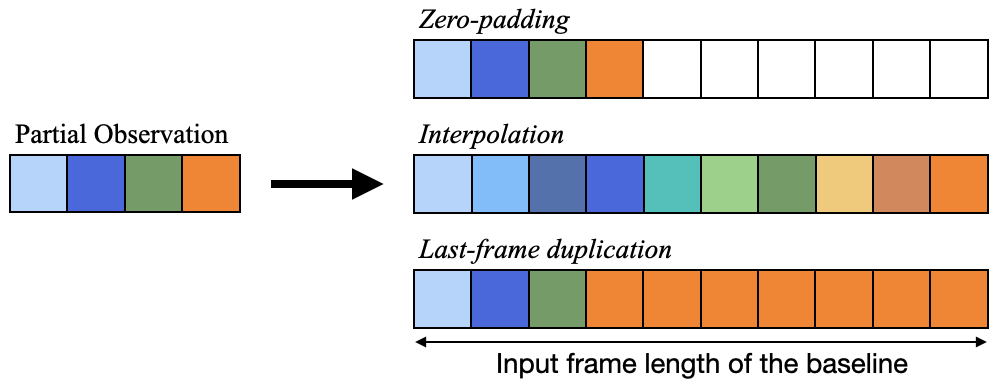}
    \caption{Visualization of different input modification strategies. Different colors represent different temporal positions. The white color indicates the zero-value padding.}
    \label{fig:input}
\end{figure}

\subsubsection{NW-UCLA~\cite{wang2014cross}} This dataset consists of 1,494 videos spanning 10 action categories, captured from three different camera perspectives. We used videos from the first two cameras for training, and those from the third camera for testing, adhering to the original paper's protocol.

\noindent \textbf{Preprocessing.}
We adhere to the methods used in previous works~\cite{chen2021channel, chi2022infogcn} for training our InfoGCN++ model. We performed normalization and localization on the skeleton data, and truncated the input video length to 64 frames for NTU RGB+D 60 and 120, and to 52 frames for NW-UCLA.

\subsection{Implementation Details}
\subsubsection{InfoGCN++}
Our model is implemented with PyTorch \cite{paszke2017automatic}, and all experiments are conducted on an Nvidia Titan RTX GPU.
To build InfoGCN++, we stack 4 encoding layers for the encoder ($L\!=\!4$) with hidden dimension 128 ($D\!=\!128$).
We train our model to predict 3 future frames ($N\!=\!3$).
We use an SGD optimizer with an initial learning rate of 0.1 and decaying the learning rate by a factor of 0.1 at the 50 and 60 epochs.
The max epoch for training is set to 70. Weight decay is set to 0.0003.
The loss coefficients $\lambda_1\!=\!1\mathrm{e}{-1}$ and $\lambda_2\!=\!1\mathrm{e}{-3}$ are used for NW-UCLA dataset and NTU RGB+D datasets.
We compare the performance of the different combinations of coefficients in \Cref{table:coefficient} and choose the one that shows the best performance.
For the NTU RGB+D datasets \cite{liu2019ntu, shahroudy2016ntu}, the batch size is set as 64.
For NW-UCLA \cite{wang2014cross}, we use batch size 32.
We publicize our code (\url{https://github.com/stnoah1/sode}) to reproduce the experimental results of InfoGCN++. 
This code includes details for data acquisition, preprocessing, environmental setup, and execution commands for the experiments in our main paper.
\subsubsection{Baseline setup}
When pretrained weights are unavailable (e.g., ST-GCN on NTU RGB+D 120), we train baseline models. 
To implement baselines, we use the code provided by the authors of the original works (ST-GCN\footnote{\url{https://github.com/yysijie/st-gcn}}, 2S-AGCN\footnote{\url{https://github.com/lshiwjx/2s-AGCN}}, MS-G3D\footnote{\url{https://github.com/kenziyuliu/MS-G3D}}, CTR-GCN\footnote{\url{https://github.com/Uason-Chen/CTR-GCN}}, InfoGCN\footnote{\url{https://github.com/stnoah1/infogcn}}).
We use pretrained weights if available; otherwise, we train the networks with the parameter provided by the authors and finetune the parameter until the performance converges.
For NTU RGB+D 60 dataset, all the baselines provide hyperparameters or pretrained weights, so we use them.
However, for NTU RGB+D 120 and NW-UCLA datasets, we train ST-GCN and 2S-AGCN using the same hyperparameter on the NTU-RGB+D 60 dataset and finetune.
We further train MS-G3D on NW-UCLA with the same strategy.

\subsection{Quantitative Results}

\subsubsection{Comparison with offline skeleton-based action recognition methods}
We contrast InfoGCN++ with offline skeleton-based action recognition methods. These include pre-trained models and those individually trained on different observation ratios. The baselines include five state-of-the-art methods: ST-GCN \cite{yan2018spatial}, 2s-AGCN \cite{shi2019two}, MS-G3D \cite{liu2020disentangling}, CTR-GCN \cite{chen2021channel}, and InfoGCN \cite{chi2022infogcn}. We use the code and pretrained weights from the original papers, modifying the input during inference for partial observations.

To assess pretrained offline skeleton-based action recognition methods, we test three input sequence modification strategies: zero-padding for future frames, interpolation of partial observation, and duplicating the last frame as per \cite{mao2019learning} (see \cref{fig:input}). The last frame duplication strategy performs best, so we used it to report baseline performance.

In Table \ref{table:sota}, we evaluate the baseline models at intervals of 20\% of the observation ratio. Our findings show that InfoGCN++ excels over all other baseline methods when the observation ratio is under 80\%. Although our model's performance matches the baselines when the observation ratio is over 80\%, it isn't the best. This trend is consistent across all datasets. Nevertheless, InfoGCN++ achieves the highest AUC on all three datasets, outdoing the baselines considerably. Notably, our performance advantage over the baselines is more pronounced at lower observation ratios. For example, at 20\% observation on the NTU 60 X-Sub, InfoGCN++ achieves 44.57\% accuracy, compared to around 10\% for other baselines. This underscores InfoGCN++'s prowess in accurately classifying actions by predicting the future, even with limited observations.

In the second part of Table \ref{table:sota}, we contrast InfoGCN++ with baselines individually trained on each partial observation. Despite being a single model capable of classifying action classes at any observation ratio, InfoGCN++'s performance holds up against these individually trained models.

\begin{figure*}[t]
    \centering
    \includegraphics[width=0.32\linewidth]{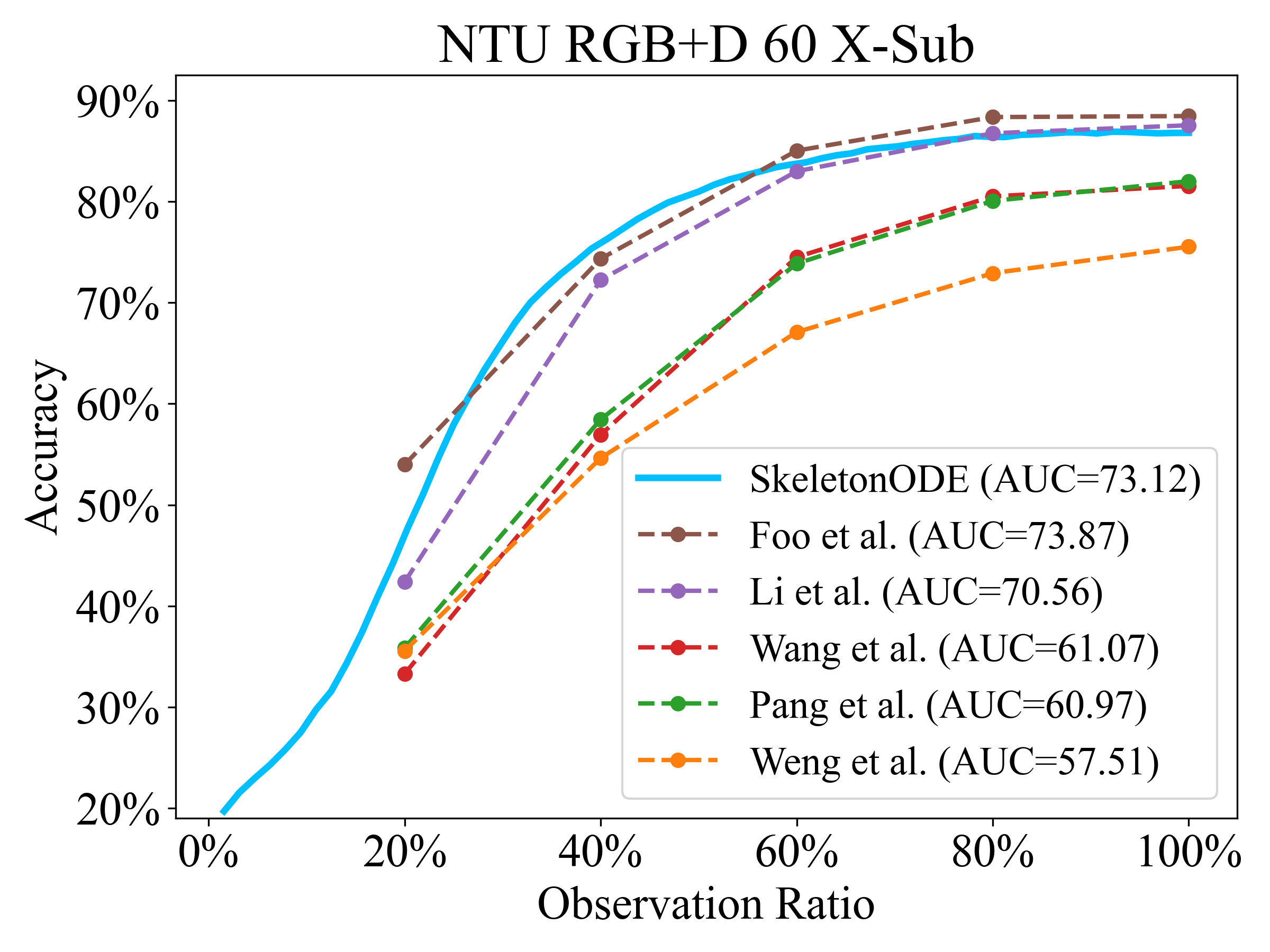}
    \includegraphics[width=0.32\linewidth]{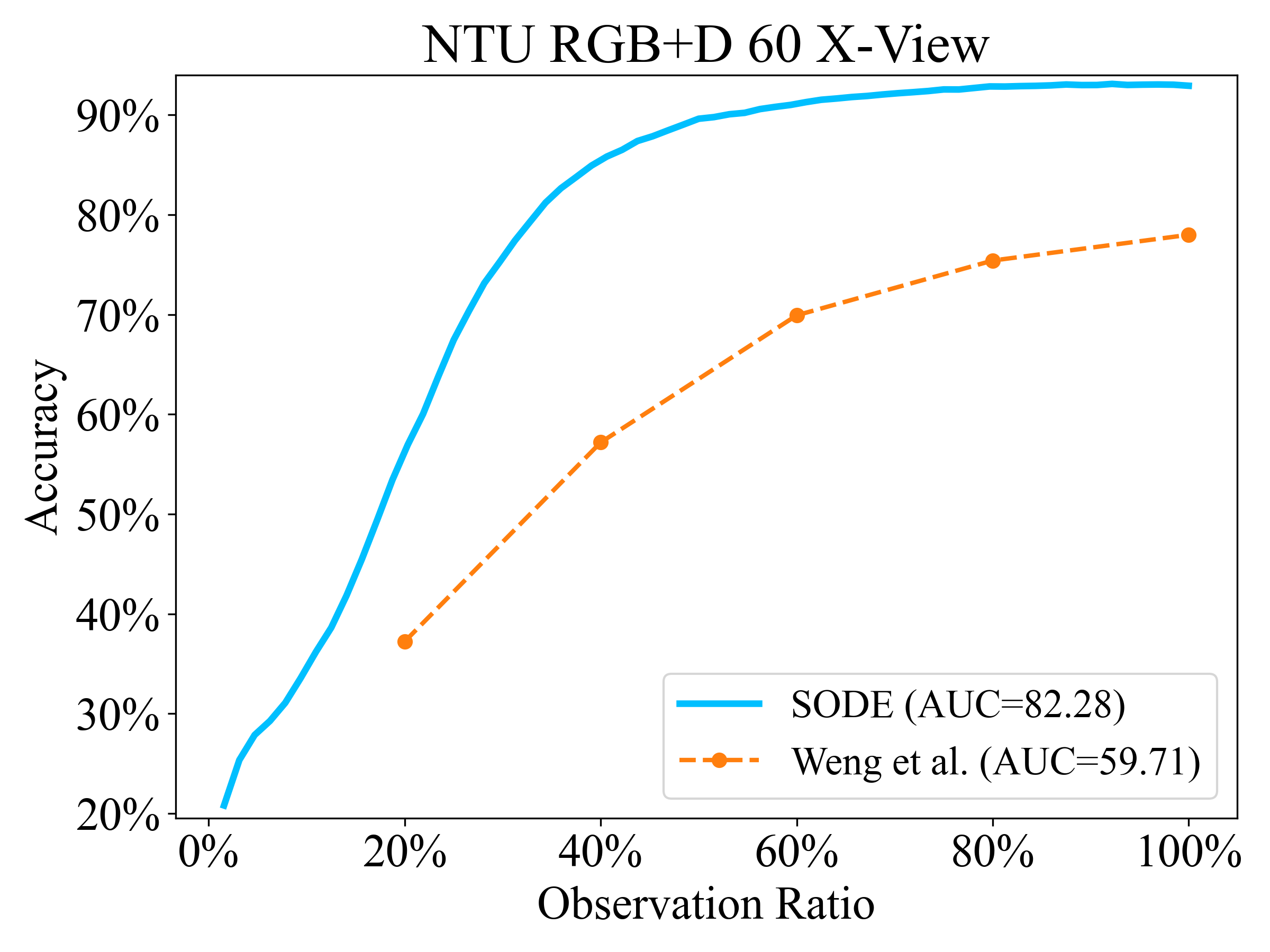}
    \includegraphics[width=0.32\linewidth]{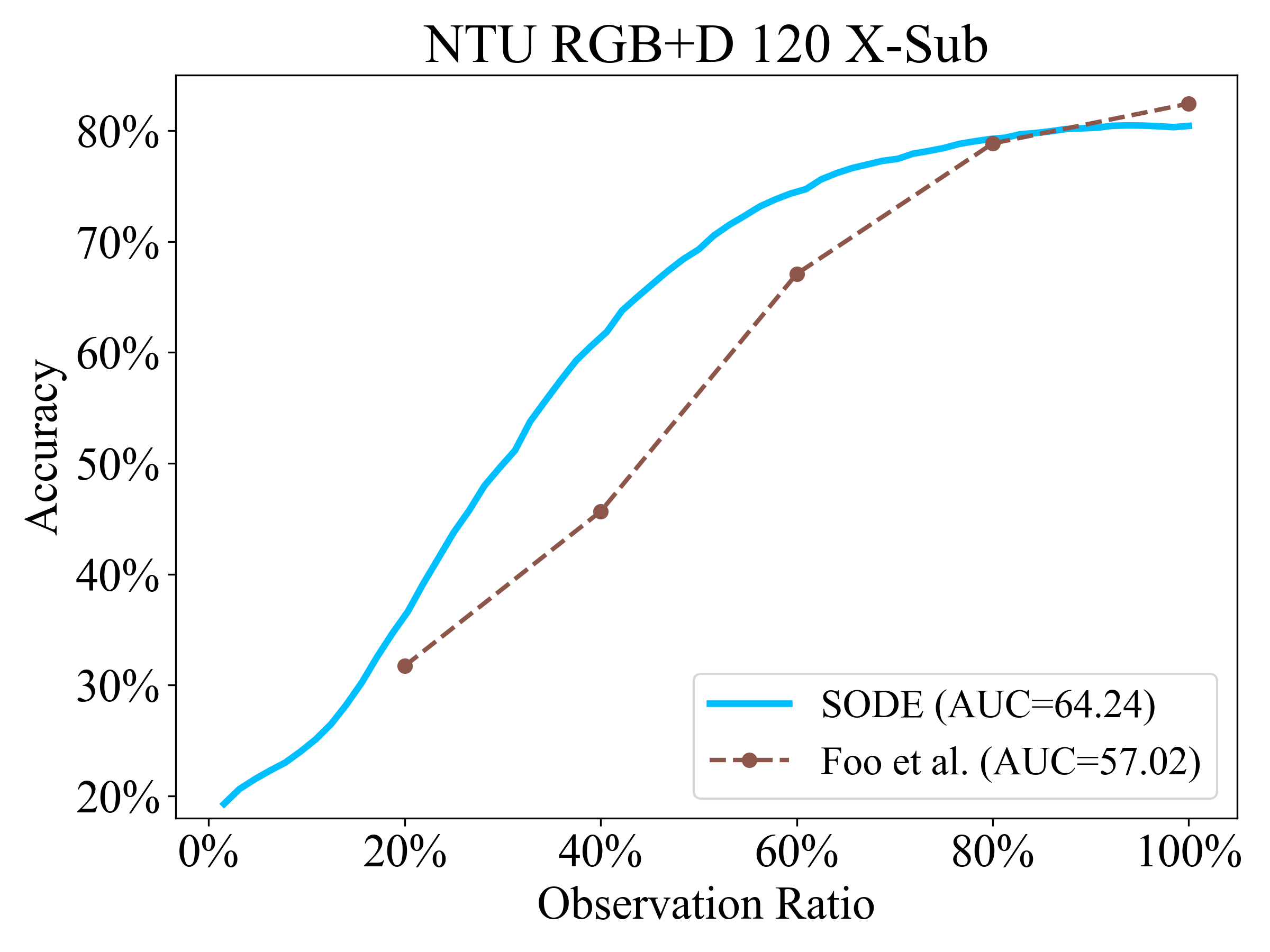}
    \caption{Performance comparison of InfoGCN++ with early action prediction methods at different observation ratios. While InfoGCN++ provides action classification in every single frame in a continuous manner, early action prediction methods produce output at a specific fraction of the observation length.
    }
    \label{fig:performance}
\end{figure*}

\begin{table*}[t]
\centering
\caption{Comparison with early action prediction methods.}
    \begin{tabular}{c l c c c c c c c}
        \hline
        \multirow{2}{*}{Datasets} & \multirow{2}{*}{Methods} & \multicolumn{5}{c}{Observation Ratio} & \multirow{2}{*}{AUC} \\
        \cline{3-7}
         &  &  20\% & 40\% & 60\% & 80\% & 100\% & & \\
        \hline\hline
        \multirow{8}{*}{NTU 60 X-Sub} & Weng \textit{et al.} \cite{weng2020early} & 35.56 & 54.63 & 67.08 & 72.91 & 75.53 & 57.51 \\
        & Wang \textit{et al.} \cite{wang2019progressive} & 35.85 & 58.45 & 73.86 & 80.06 & 82.01 & 60.97  \\
        & Pang \textit{et al.} \cite{pang2019dbdnet} & 33.30 & 56.94 & 74.50 & 80.51 & 81.54 & 61.07  \\
        & Wang \textit{et al.} \cite{wang2021dear} & 32.72 & 69.71 & 80.18 & 83.49 & 84.45 & - \\
        & Li \textit{et al.} \cite{li2020hard} & 42.39 & 72.24 & 82.99 & 86.75 & 87.54 & 70.56  \\
        & Foo \textit{et al.} \cite{foo2022era} & \textbf{53.98} & 74.34 & \textbf{85.03} & \textbf{88.35} & \textbf{88.45} & \textbf{73.87} \\
        \cline{2-9}
        & \textbf{InfoGCN++ (Ours)} & 47.66 & \textbf{76.13} & 83.91 & 86.47 & 87.02 & 73.14 \\
        \hline
        \multirow{2}{*}{NTU 60 X-View} & Weng \textit{et al.} \cite{weng2020early} & 37.22 & 57.18 & 69.92 & 75.41 & 77.99 & 59.71  \\
        \cline{2-9}
        & \textbf{InfoGCN++ (Ours)} & \textbf{59.11} & \textbf{87.48} & \textbf{92.56} & \textbf{93.89} & \textbf{93.92} & \textbf{82.28} \\
        \hline
        \multirow{2}{*}{NTU 120 X-Sub} & Foo \textit{et al.} \cite{foo2022era} & 31.73 & 45.67 & 67.08 & 78.84 & \textbf{82.43} & 57.02  \\
        \cline{2-9}
        & \textbf{InfoGCN++ (Ours)}  & \textbf{37.28} & \textbf{62.81} & \textbf{75.43} & \textbf{79.67} & 80.65 & \textbf{64.24} \\
        \hline
        \multirow{2}{*}{NTU 120 X-Set} & \multirow{2}{*}{\textbf{InfoGCN++ (Ours)}} & \multirow{2}{*}{\textbf{41.03}} & \multirow{2}{*}{\textbf{67.60}} & \multirow{2}{*}{\textbf{78.54}} & \multirow{2}{*}{\textbf{82.47}} & \multirow{2}{*}{\textbf{83.22}} & \multirow{2}{*}{\textbf{67.69}}  \\
        \\
        \hline
        \begin{tabular}[c]{@{}c@{}}NW-UCLA\end{tabular} & {\textbf{InfoGCN++ (Ours)}} & \textbf{78.02} & \textbf{85.99} & \textbf{88.15} & \textbf{90.73} & \textbf{90.09} & \textbf{85.47} &\\
        \hline
    \end{tabular}
\label{table:EAR}
\end{table*}

\subsubsection{Comparison with early prediction methods}
\cref{fig:performance} and \Cref{table:EAR} showcases the superior performance of InfoGCN++ over previous early action prediction methods. We adopt the two-stream setup from Foo \textit{et al.} \cite{foo2022era} to ensure an unbiased comparison. Our model is assessed by calculating classification accuracy at intervals of 20\% observation ratios, reporting the area under the curve (AUC) as a single metric, in line with \cite{foo2022era, li2020hard}. Our results on the NTU 120 X-Set and NW-UCLA datasets are also included to spur further research. Despite being created for continuous action classification, InfoGCN++ demonstrates versatility, performing comparably or better than early action prediction methods. Specifically, we match the AUC on the NTU 60 X-Sub split, and surpass the state-of-the-art method on the NTU 60 X-Sub and NTU-120 X-Set datasets. Unlike early action prediction methods that only output at specific observation fractions, InfoGCN++ provides dense inference. Additionally, with an inference speed of 18ms per frame, InfoGCN++ is apt for real-time applications.

\begin{table}[t]
    \centering
    \caption{Comparison on different loss setups.}
    \addtolength{\tabcolsep}{-1pt}    
    \begin{tabular}{c c c c c c c c c c}
        \hline
        \multicolumn{3}{c}{Losses} & \multicolumn{5}{c}{Observation Ratio} & \multirow{2}{*}{AUC}\\         \cline{4-8}
        $\mathcal{L}_{\text{cls}}$ & $\mathcal{L}_{\text{pred}}$ & $\mathcal{L}_{\text{feat}}$ & 20\% & 40\% & 60\% & 80\% & 100\% & \\
        \hline\hline
        \checkmark & & & 73.28& 80.82 & 83.19 & 86.21 & 86.85 & 81.83\\
        \checkmark &\checkmark& & 77.80 & 84.48 & 87.93 & 89.22 & 89.44 & 84.85\\
        \checkmark &\checkmark&\checkmark&  \textbf{78.02}& \textbf{85.99}& \textbf{88.15}& \textbf{90.73}& \textbf{90.09}& \textbf{85.47}\\
        \hline
    \end{tabular}
    \label{table:loss}
\end{table}

\subsection{Ablation Studies and Analysis}
\noindent This section features ablation studies and analyses, investigating the effects of individual design components within InfoGCN++. Unless otherwise stated, all studies are conducted on the NW-UCLA \cite{wang2014cross} dataset.

\begin{figure}[t]
  \centering
  \includegraphics[width=1.0\linewidth]{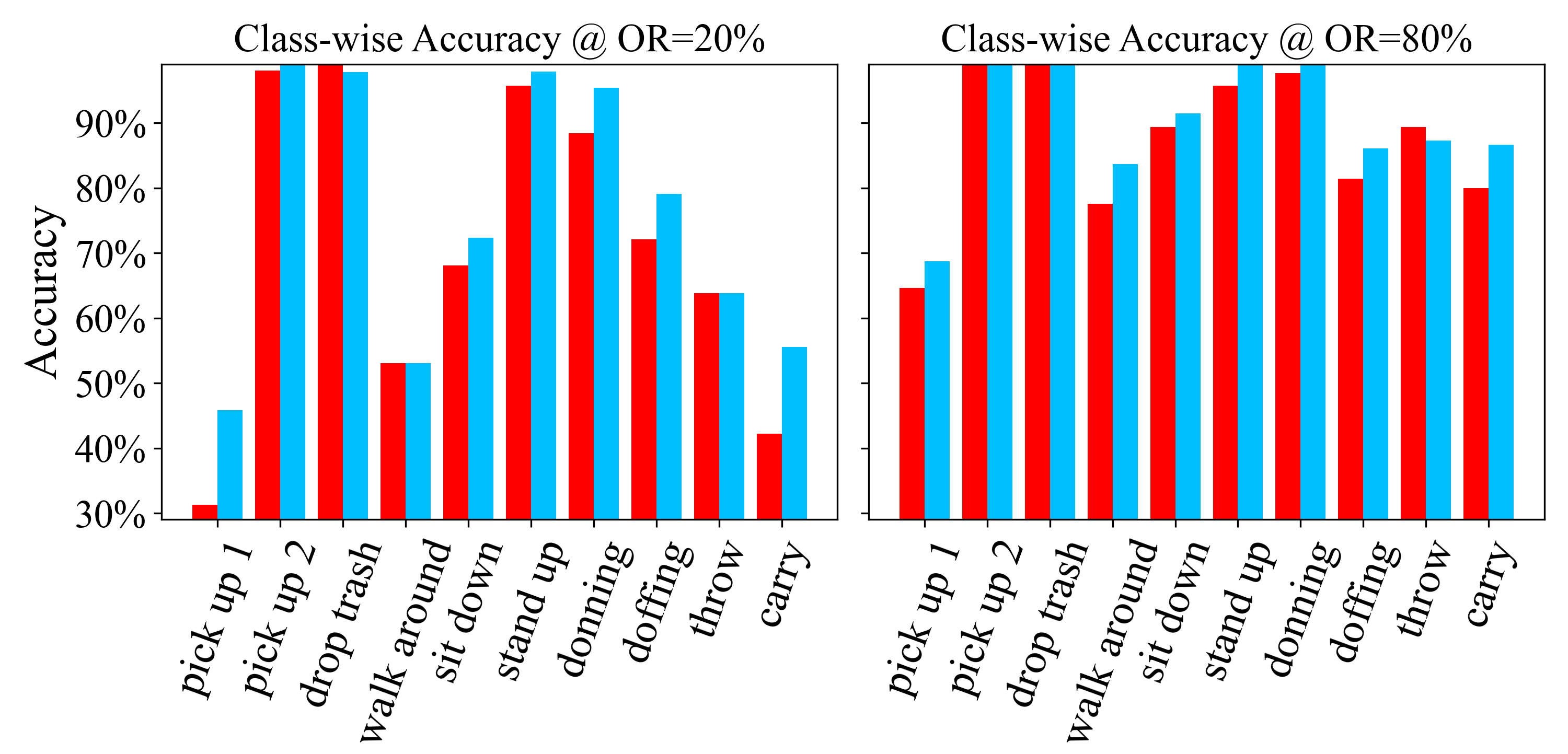}
  \caption{Class-wise performance comparison of the InfoGCN++ with and without future motion prediction at 20\% and 80\% observation ratio (OR), where the red and blue colors correspond to the result of the first and third rows in \Cref{table:loss}, respectively.}
  \label{fig:class accuracy}
\end{figure}

\begin{table}[t]
\centering
    \caption{Comparative results of future motion prediction methods.}
    \begin{tabular}{l c c}
        \hline
        \multirow{2}{*}{Future Predictor} &\multirow{2}{*}{AUC $\uparrow$} & \multirow{2}{*}{$\mathcal{L}_{\text{rec}}$ $\downarrow$} \\ 
        && \\
        \hline\hline
        Baseline & 81.83 & -\\
        RNN & 83.38 & 0.0302\\ \hline
        NeuralODE~\cite{chen2018neural} &\textbf{85.47} & \textbf{0.0092}\\
        \hline
    \end{tabular}
    \label{table:numerical}
\end{table}

\begin{table}[t]
\centering
\caption{Comparison on different numbers of prediction steps.}
\resizebox{1.0\linewidth}{!}{
    \begin{tabular}{c c c c c c c c c}
        \hline
        \multirow{2}{*}{\begin{tabular}[c]{@{}c@{}}\# Prediction Step \\ (N)\end{tabular}} & \multicolumn{5}{c}{Observation Ratio} & \multirow{2}{*}{AUC}\\
        \cmidrule(lr){2-6}
        & 20\% & 40\% & 60\% & 80\% & 100\% & \\
        \hline\hline
        0 & 73.28 & 80.82 & 83.19 & 86.21 & 86.85 & 81.47\\
        1 & 74.35 & 82.33 & 85.13 & 85.99 & 87.50 & 82.00\\
        2 & 73.71 & 83.41 & 85.13 & 87.28 & 88.36 & 82.61\\
        3 & \textbf{78.02}& \textbf{85.99}& 88.15 & \textbf{90.73}& \textbf{90.09}& \textbf{85.47}\\
        4 & 75.86 & 83.62 & 85.13 & 88.15 & 89.01 & 83.64\\
        5 & 76.51 & 84.48 & \textbf{88.36} & 89.98 & \textbf{90.09} & 84.78\\
        \hline
    \end{tabular}
   }    
    \vspace{-.3em}
    \label{table:n_step}
\end{table}

\begin{table}[!t]
    \centering
        \caption{Comparative results of different encoder configurations.}
    \begin{tabular}{l c c}
        \hline
        Encoders & \# Params & AUC\\
        \hline\hline
        RNN & 2.07M & 77.27 \\ \hline
        Transformer&\\
        \hspace{.5em} w/ linear & 1.47M & 81.18 \\
        \hspace{.5em} w/ GCN & 2.04M & 83.23\\
        \hspace{.5em} w/ SA-GC \cite{chi2022infogcn} & 2.09M  & \textbf{85.47}\\
        \hline
    \end{tabular}
    \label{table:backbone}
\end{table}

\begin{table}[t]
\centering
\caption{AUC comparison between different loss coefficients on NW-UCLA dataset.}
\begin{tabular}{|l l| ccc|}
\hline
\multicolumn{2}{|l|}{}                            & \multicolumn{3}{c|}{$\lambda_1$}                                \\ \cline{3-5} 
\multicolumn{2}{|l|}{\multirow{-2}{*}{}}          & $1\mathrm{e}{-2}$ & $1\mathrm{e}{-1}$ & $1$          \\ \hline
\multicolumn{1}{|l|}{}                              & $1\mathrm{e}{-3}$          & 84.80                & \textbf{85.47}               & 83.80          \\
\multicolumn{1}{|l|}{\multirow{-1}{*}{$\lambda_2$}}                              & $1\mathrm{e}{-2}$          & 83.77                & 84.62               & 84.21 \\
\multicolumn{1}{|l|}{}                              & $1\mathrm{e}{-1}$          & 82.96                & 83.36               & 84.57          \\\hline
\end{tabular}
\label{table:coefficient}
\end{table}

\begin{table}[t]
\centering
\caption{The inference speed of InfoGCN++ on different datasets.}
\begin{tabular}{lc}
\hline
\multicolumn{1}{c}{Datasets} & \multicolumn{1}{c}{Inference Speed} \\
\hline\hline
NTU RGB+D 60 \& 120   & 18.0 ms (56 fps) \\ \hline
NW-UCLA      & 15.6 ms (64 fps) \\ \hline
\end{tabular}
\label{table:inference_speed}
\end{table}

\subsubsection{Contribution of Future Motion Prediction}
Our first line of inquiry pertains to the role of future motion prediction in online skeleton-based action recognition. In Table \ref{table:loss}, we compare our model's performance using different loss setups when predicting future 3 time steps ($N=3$) motion. The model trained solely with $\mathcal{L}_{\text{cls}}$ (i.e., without future motion prediction) forms the baseline. Our findings reveal that the model's AUC increases by 3.17 upon adding $\mathcal{L}_{\text{rec}}$ to $\mathcal{L}_{\text{cls}}$, and by an additional 1.36 upon incorporating $\mathcal{L}_{\text{feat}}$. This highlights the significant contribution of future motion prediction to online skeleton-based action recognition. In \cref{fig:class accuracy}, we display class-wise accuracy at different observation ratios (OR), showing that future motion prediction enhances classification accuracy across almost all action categories at both 20\% and 80\% OR. Specifically, complex actions like \textit{pick up 1'} and \textit{carry} that are challenging to discern early in motion and exhibit low accuracy at 20\% observation, witness a substantial accuracy boost when employing InfoGCN++'s future motion prediction.

\subsubsection{Advantage of NeuralODE in Future Motion Prediction}
Next, we demonstrate the edge of NeuralODE over RNN in predicting future motion. Table \ref{table:numerical} presents a comparison between different future motion prediction methods, including the baseline InfoGCN++ without future motion prediction. In our RNN construct, we modify Conv-LSTM \cite{shi2015convolutional} by substituting convolution with the graph convolution of GCN (\cref{eq:gcn}) and stacking two RNN layers. The NeuralODE outperforms the RNN, yielding lower prediction loss and higher AUC, thereby echoing previous works \cite{chen2018neural, rubanova2019latent} that have championed NeuralODE's capacity to model sequential data and predict future steps with observed values and captured physics.

\begin{figure*}[t]
    \centering
    \includegraphics[width=1.0\linewidth]{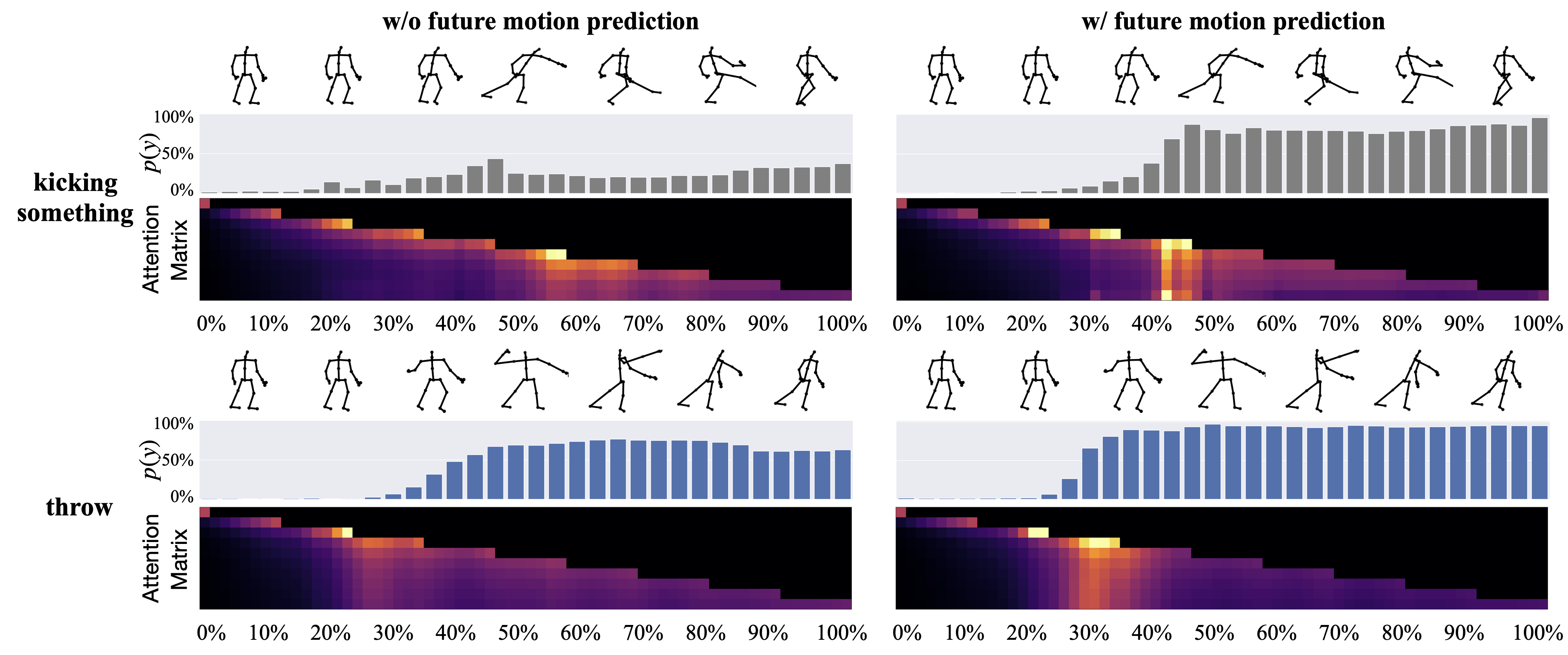}
    \caption{Qualitative samples from NTU 120 X-Sub. (\textbf{left}) without future prediction and (\textbf{right}) with future prediction of $N = 3$. For attention visualization, we randomly select one head from the last encoding layer and average the value over spatial dimension. The brightness in the attention map indicates the magnitude of attention. We uniformly sample the rows of attention maps for simplicity.}
    \label{fig:qualitative}
\end{figure*}

\subsubsection{The Number of Future Prediction Steps}
In Table \ref{table:n_step}, we delve into how the number of future motion prediction steps influences performance. We discern that the model performs optimally at $N=3$, equivalent to 5\% of the entire sequence. This implies that lengthier prediction steps during training do not necessarily yield a better representation. We posit that this is due to the increased uncertainty when predicting the long-term future, which could impede the learning of a discriminative representation for action classification.

\subsubsection{Encoder Configurations}
Table \ref{table:backbone} compares different encoder designs. Transformer-based encoders outperform RNN-based ones significantly, even when bearing similar or fewer parameters, thereby establishing the transformer's superiority in temporal modeling. Moreover, the Transformer coupled with GCN projections offers the highest AUC, underlining the advantage of our skeleton feature aggregation method in amalgamating spatio-temporal information. Furthermore, the Transformer with SA-GC by \cite{chi2022infogcn} outperforms the standard GCN, suggesting that InfoGCN++ capturing the intrinsic topology of the skeleton proves beneficial in online skeleton-based action recognition.

\subsubsection{Coefficients for Multi-task Learning}
We performed a grid search over the loss coefficients space, specifically in the range of $\lambda_1 \in \{1\mathrm{e}{-2}, 1\mathrm{e}{-1}, 1\}$, and $ \lambda_2 \in \{1\mathrm{e}{-3}, 1\mathrm{e}{-2}, 1\mathrm{e}{-1}\}$.The resulting performance across various combinations is summarized in \Cref{table:coefficient}. We select the coefficient pair that yielded the best performance to be utilized in our experiments.

\subsubsection{Inference Speed}
Finally, we evaluated InfoGCN++'s inference speed on both the NTU RGB+D and NW-UCLA datasets, as reported in Table \ref{table:inference_speed}. InfoGCN++ exhibits an average inference speed of 18.0 ms (equivalent to 56 fps) for the NTU RGB+D dataset, and 15.6 ms (or 64 fps) for the NW-UCLA dataset. This finding showcases InfoGCN++'s capability for real-time inference—an essential requirement for diverse practical applications. Since the NTU RGB+D 60 and 120 datasets share the same graph topology, we present their inference speeds as a single value.

\subsubsection{Attention Visualization}
\begin{figure*}[!htpb]
    \centering
    \includegraphics[width=0.95\linewidth]{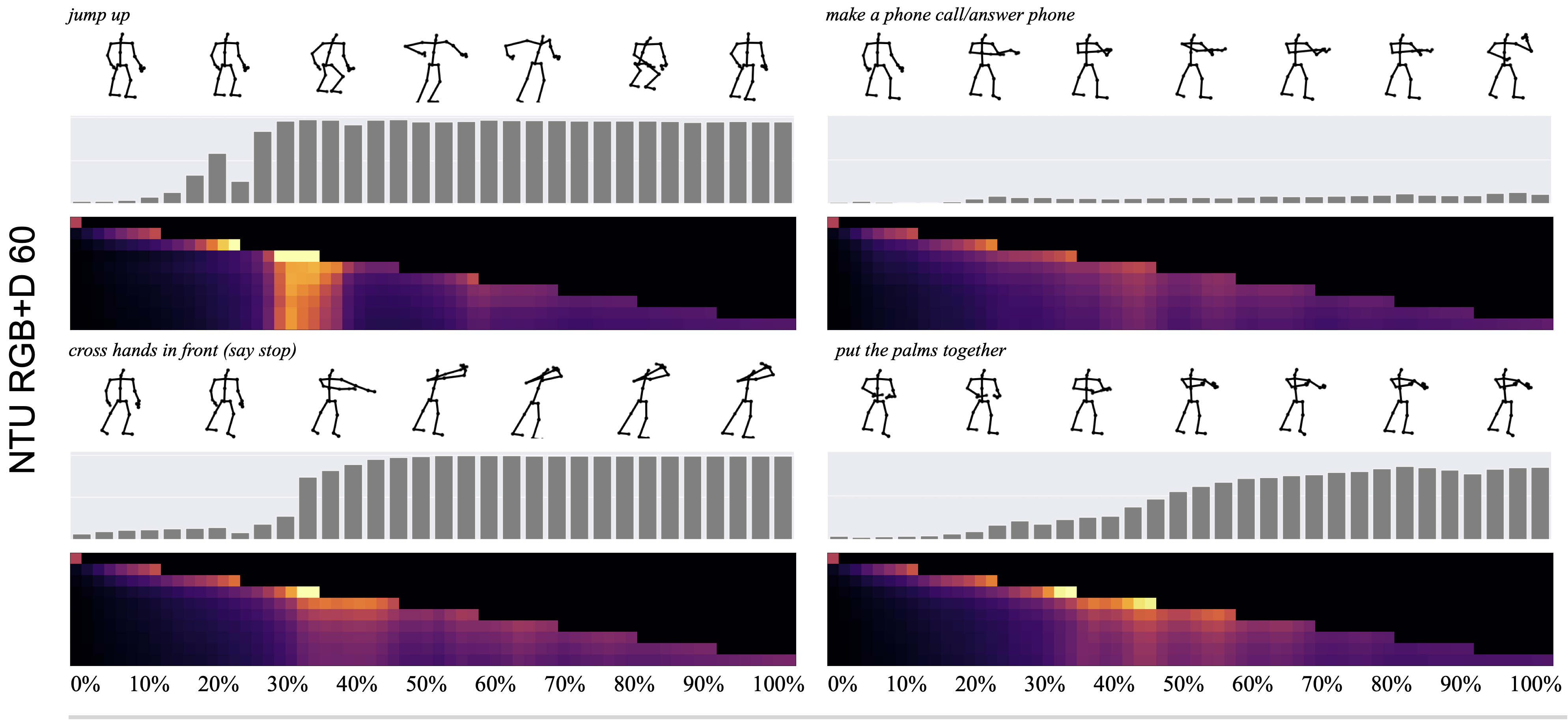}
    \includegraphics[width=0.95\linewidth]{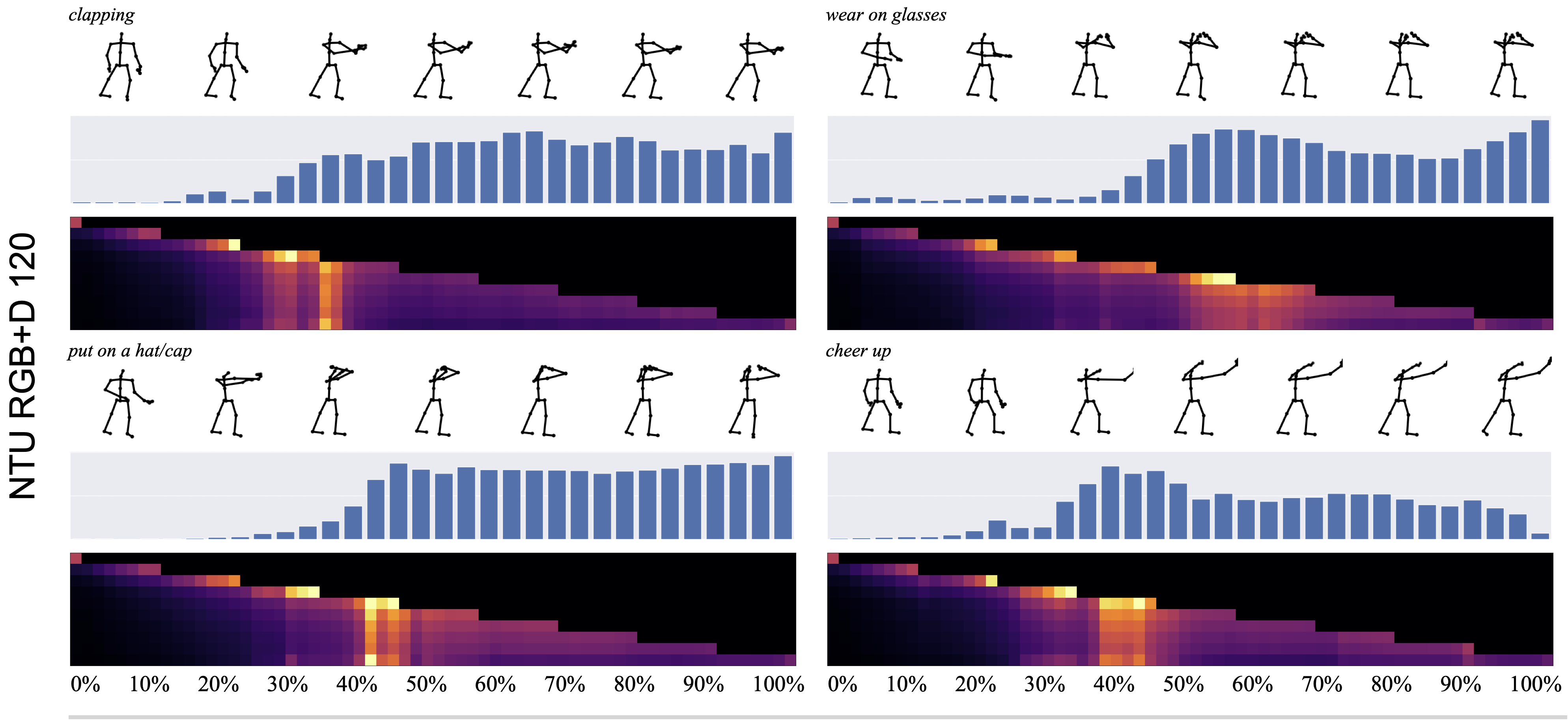}
    \includegraphics[width=0.95\linewidth]{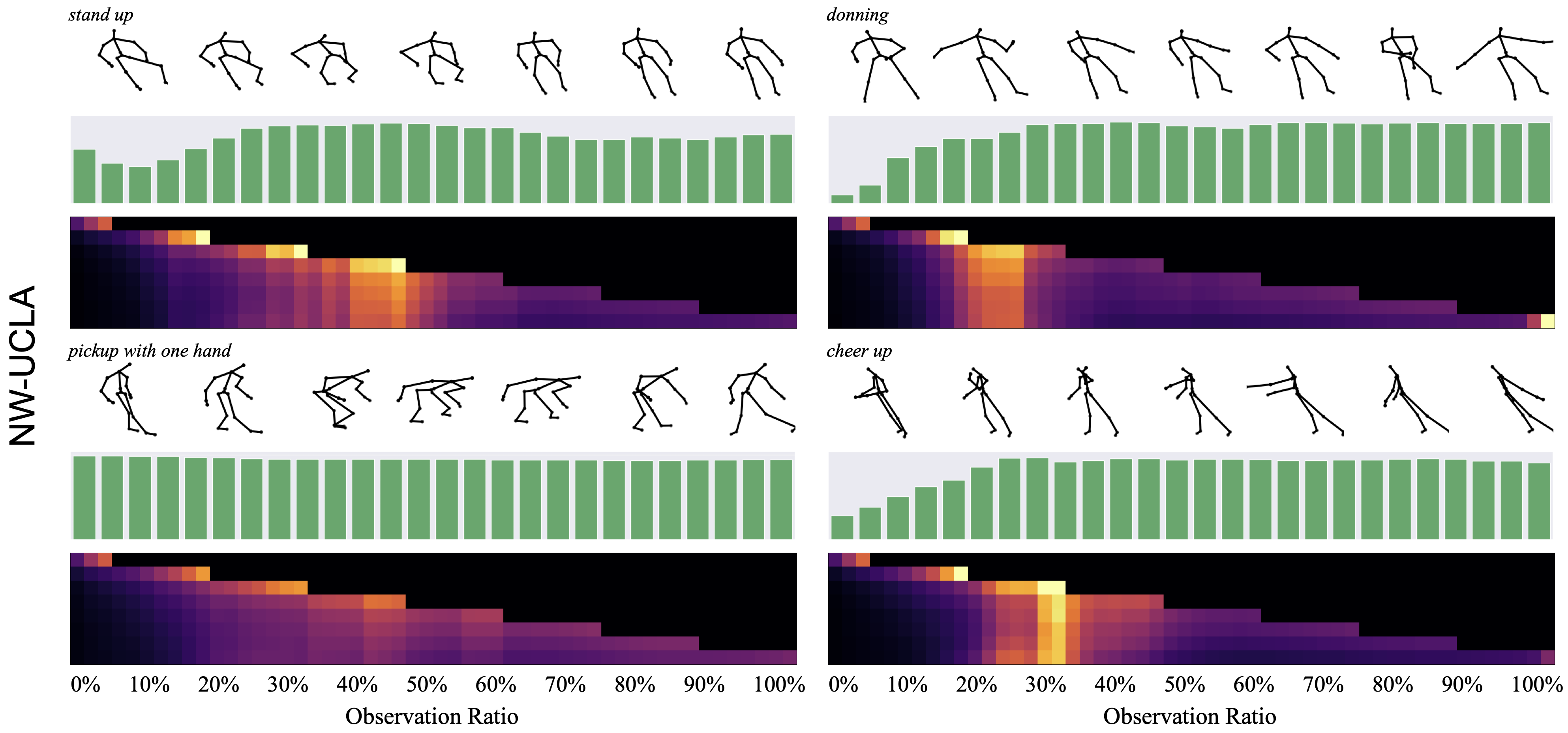}
    \caption{\textbf{Qualitative Results}. We provide a corresponding classification score and attention matrix similar to \cref{fig:qualitative}.}
    \label{fig:qualitative2}
\end{figure*}

\cref{fig:qualitative} presents a comparison of the attention maps produced by InfoGCN++'s encoder for models trained with and without future motion prediction. It's observed that the model trained with future motion prediction shows a marked correlation between the attention map and prediction probability, with distinct frames receiving more attention than others. Conversely, the model trained without future motion prediction demonstrates a tendency to focus more on recent time frames rather than on specific ones. This observation implies that future motion prediction guides the encoder to discern which frames to focus on in order to accurately recognize the action from partial observations. Additional examples from other datasets are provided in \cref{fig:qualitative2}.

\subsection{Qualitative Results}
\noindent We further illustrate qualitative results derived from the NTU RGB+D 60, NTU RGB+D 120, and NW-UCLA datasets in \cref{fig:qualitative2}. We depict the outcomes from the InfoGCN++ trained with a future prediction step of $N=3$. Each row in the figure shows the skeleton sequence, the classification score ($p(y)$), and the attention map, mirroring the format of figures in the main paper. The attention maps exhibit similar patterns to those presented in the main paper. The model prioritizes frames with discriminative motion within the input sequence, and a notable increase in the classification score $p(y)$ is observed around these attended frames.

%% file: sections/5_conclusion.tex
\section{Discussion}

\subsection{Length of Action Sequence}
\noindent Our proposed InfoGCN++ framework encodes representation based on all previous observations. However, this approach encounters a limitation concerning the memory requirement. Specifically, memory usage grows quadratically with an increase in sequence length, rendering InfoGCN++ less scalable for longer-term sequences. This constraint offers a direction for our future work, where we aim to develop a memory-efficient design of InfoGCN++ that is capable of handling extended action sequences.

\subsection{Model Performance on Full Observation}
\noindent In terms of performance across the entire observation of the sequence (i.e., 100\% observation ratio), InfoGCN++ delivers results that are comparable to, but not superior to, current skeleton-based action recognition algorithms (i.e. InfoGCN~\cite{chi2022infogcn}). We envision integrating the strengths of other methods into our framework to enhance its performance at full observation. For example, incorporating intrinsic topology modeling in our encoder may help push InfoGCN++'s performance closer to the state-of-the-art. This aspiration forms part of our agenda for future enhancements to the system.

\section{Conclusion}
\noindent In this work, we introduced InfoGCN++, an extended version of InfoGCN for online skeleton-based action recognition, which integrates the strengths of Neural ODEs with Transformer encoders to simultaneously predict and classify actions in real-time. Our method showed superior performance in comparison to other early action prediction methods and offline skeleton-based action recognition methods, highlighting its versatility. Through ablation studies, we validated the significance of future motion prediction and the advantage of Neural ODEs over RNNs. Furthermore, we demonstrated that our model exhibits robust performance across multiple datasets and offers a valuable tool for real-time applications due to its rapid inference speed. We also highlighted the potential of future motion prediction in guiding the encoder to focus on the most informative frames. Our results not only substantiate the efficacy of InfoGCN++ but also open new avenues for future research in online action recognition and prediction.

\section{Acknowledgments}
\noindent This work was partially supported by US National Science Foundation (FW-HTF 1839971). We also acknowledge the Feddersen Chair Funds for Professor Karthik Ramani.


%% file: main.bbl
\begin{thebibliography}{10}
\providecommand{\url}[1]{#1}
\csname url@samestyle\endcsname
\providecommand{\newblock}{\relax}
\providecommand{\bibinfo}[2]{#2}
\providecommand{\BIBentrySTDinterwordspacing}{\spaceskip=0pt\relax}
\providecommand{\BIBentryALTinterwordstretchfactor}{4}
\providecommand{\BIBentryALTinterwordspacing}{\spaceskip=\fontdimen2\font plus
\BIBentryALTinterwordstretchfactor\fontdimen3\font minus
  \fontdimen4\font\relax}
\providecommand{\BIBforeignlanguage}[2]{{%
\expandafter\ifx\csname l@#1\endcsname\relax
\typeout{** WARNING: IEEEtran.bst: No hyphenation pattern has been}%
\typeout{** loaded for the language `#1'. Using the pattern for}%
\typeout{** the default language instead.}%
\else
\language=\csname l@#1\endcsname
\fi
#2}}
\providecommand{\BIBdecl}{\relax}
\BIBdecl

\bibitem{wang2021gesturar}
T.~Wang, X.~Qian, F.~He, X.~Hu, Y.~Cao, and K.~Ramani, ``Gesturar: An authoring
  system for creating freehand interactive augmented reality applications,'' in
  \emph{The 34th Annual ACM Symposium on User Interface Software and
  Technology}, 2021, pp. 552--567.

\bibitem{huang2021adaptutar}
G.~Huang, X.~Qian, T.~Wang, F.~Patel, M.~Sreeram, Y.~Cao, K.~Ramani, and A.~J.
  Quinn, ``Adaptutar: An adaptive tutoring system for machine tasks in
  augmented reality,'' in \emph{Proceedings of the 2021 CHI Conference on Human
  Factors in Computing Systems}, 2021, pp. 1--15.

\bibitem{pei2011parsing}
M.~Pei, Y.~Jia, and S.-C. Zhu, ``Parsing video events with goal inference and
  intent prediction,'' in \emph{2011 International Conference on Computer
  Vision}.\hskip 1em plus 0.5em minus 0.4em\relax IEEE, 2011, pp. 487--494.

\bibitem{vondrick2016anticipating}
C.~Vondrick, H.~Pirsiavash, and A.~Torralba, ``Anticipating visual
  representations from unlabeled video,'' in \emph{Proceedings of the IEEE
  conference on computer vision and pattern recognition}, 2016, pp. 98--106.

\bibitem{brehar2021pedestrian}
R.~D. Brehar, M.~P. Muresan, T.~Mari{\c{t}}a, C.-C. Vancea, M.~Negru, and
  S.~Nedevschi, ``Pedestrian street-cross action recognition in monocular far
  infrared sequences,'' \emph{IEEE Access}, vol.~9, pp. 74\,302--74\,324, 2021.

\bibitem{pop2019multi}
D.~O. Pop, A.~Rogozan, C.~Chatelain, F.~Nashashibi, and A.~Bensrhair,
  ``Multi-task deep learning for pedestrian detection, action recognition and
  time to cross prediction,'' \emph{IEEE Access}, vol.~7, pp.
  149\,318--149\,327, 2019.

\bibitem{liu2020disentangling}
Z.~Liu, H.~Zhang, Z.~Chen, Z.~Wang, and W.~Ouyang, ``Disentangling and unifying
  graph convolutions for skeleton-based action recognition,'' in
  \emph{Proceedings of the IEEE/CVF Conference on Computer Vision and Pattern
  Recognition}, 2020, pp. 143--152.

\bibitem{chen2021channel}
Y.~Chen, Z.~Zhang, C.~Yuan, B.~Li, Y.~Deng, and W.~Hu, ``Channel-wise topology
  refinement graph convolution for skeleton-based action recognition,'' in
  \emph{Proceedings of the IEEE/CVF International Conference on Computer
  Vision}, 2021, pp. 13\,359--13\,368.

\bibitem{yan2018spatial}
S.~Yan, Y.~Xiong, and D.~Lin, ``Spatial temporal graph convolutional networks
  for skeleton-based action recognition,'' in \emph{Thirty-second AAAI
  conference on artificial intelligence}, 2018.

\bibitem{zhang2020semantics}
P.~Zhang, C.~Lan, W.~Zeng, J.~Xing, J.~Xue, and N.~Zheng, ``Semantics-guided
  neural networks for efficient skeleton-based human action recognition,'' in
  \emph{Proceedings of the IEEE/CVF Conference on Computer Vision and Pattern
  Recognition}, 2020, pp. 1112--1121.

\bibitem{shi2019two}
L.~Shi, Y.~Zhang, J.~Cheng, and H.~Lu, ``Two-stream adaptive graph
  convolutional networks for skeleton-based action recognition,'' in
  \emph{Proceedings of the IEEE/CVF conference on computer vision and pattern
  recognition}, 2019, pp. 12\,026--12\,035.

\bibitem{shi2019skeleton}
------, ``Skeleton-based action recognition with directed graph neural
  networks,'' in \emph{Proceedings of the IEEE/CVF Conference on Computer
  Vision and Pattern Recognition}, 2019, pp. 7912--7921.

\bibitem{li2019actional}
M.~Li, S.~Chen, X.~Chen, Y.~Zhang, Y.~Wang, and Q.~Tian, ``Actional-structural
  graph convolutional networks for skeleton-based action recognition,'' in
  \emph{Proceedings of the IEEE/CVF conference on computer vision and pattern
  recognition}, 2019, pp. 3595--3603.

\bibitem{cheng2020skeleton}
K.~Cheng, Y.~Zhang, X.~He, W.~Chen, J.~Cheng, and H.~Lu, ``Skeleton-based
  action recognition with shift graph convolutional network,'' in
  \emph{Proceedings of the IEEE/CVF Conference on Computer Vision and Pattern
  Recognition}, 2020, pp. 183--192.

\bibitem{korban2020ddgcn}
M.~Korban and X.~Li, ``Ddgcn: A dynamic directed graph convolutional network
  for action recognition,'' in \emph{European Conference on Computer
  Vision}.\hskip 1em plus 0.5em minus 0.4em\relax Springer, 2020, pp. 761--776.

\bibitem{ye2020dynamic}
F.~Ye, S.~Pu, Q.~Zhong, C.~Li, D.~Xie, and H.~Tang, ``Dynamic gcn:
  Context-enriched topology learning for skeleton-based action recognition,''
  in \emph{Proceedings of the 28th ACM International Conference on Multimedia},
  2020, pp. 55--63.

\bibitem{chen2021multi}
Z.~Chen, S.~Li, B.~Yang, Q.~Li, and H.~Liu, ``Multi-scale spatial temporal
  graph convolutional network for skeleton-based action recognition,'' in
  \emph{Proceedings of the AAAI conference on artificial intelligence},
  vol.~35, no.~2, 2021, pp. 1113--1122.

\bibitem{chi2022infogcn}
H.-g. Chi, M.~H. Ha, S.~Chi, S.~W. Lee, Q.~Huang, and K.~Ramani, ``Infogcn:
  Representation learning for human skeleton-based action recognition,'' in
  \emph{Proceedings of the IEEE/CVF Conference on Computer Vision and Pattern
  Recognition}, 2022, pp. 20\,186--20\,196.

\bibitem{shahroudy2016ntu}
A.~Shahroudy, J.~Liu, T.-T. Ng, and G.~Wang, ``Ntu rgb+ d: A large scale
  dataset for 3d human activity analysis,'' in \emph{Proceedings of the IEEE
  conference on computer vision and pattern recognition}, 2016, pp. 1010--1019.

\bibitem{van2020ergonomic}
M.~K. van~den Broek and T.~B. Moeslund, ``Ergonomic adaptation of robotic
  movements in human-robot collaboration,'' in \emph{Companion of the 2020
  ACM/IEEE International Conference on Human-Robot Interaction}, 2020, pp.
  499--501.

\bibitem{foo2022era}
L.~G. Foo, T.~Li, H.~Rahmani, Q.~Ke, and J.~Liu, ``Era: Expert retrieval and
  assembly for early action prediction,'' in \emph{Computer Vision--ECCV 2022:
  17th European Conference, Tel Aviv, Israel, October 23--27, 2022,
  Proceedings, Part XXXIV}.\hskip 1em plus 0.5em minus 0.4em\relax Springer,
  2022, pp. 670--688.

\bibitem{wang2021dear}
R.~Wang, J.~Liu, Q.~Ke, D.~Peng, and Y.~Lei, ``Dear-net: Learning diversities
  for skeleton-based early action recognition,'' \emph{IEEE Transactions on
  Multimedia}, 2021.

\bibitem{li2020hard}
T.~Li, J.~Liu, W.~Zhang, and L.~Duan, ``Hard-net: Hardness-aware discrimination
  network for 3d early activity prediction,'' in \emph{European Conference on
  Computer Vision}.\hskip 1em plus 0.5em minus 0.4em\relax Springer, 2020, pp.
  420--436.

\bibitem{wang2019progressive}
X.~Wang, J.-F. Hu, J.-H. Lai, J.~Zhang, and W.-S. Zheng, ``Progressive
  teacher-student learning for early action prediction,'' in \emph{Proceedings
  of the IEEE/CVF Conference on Computer Vision and Pattern Recognition}, 2019,
  pp. 3556--3565.

\bibitem{hu2018early}
J.-F. Hu, W.-S. Zheng, L.~Ma, G.~Wang, J.~Lai, and J.~Zhang, ``Early action
  prediction by soft regression,'' \emph{IEEE transactions on pattern analysis
  and machine intelligence}, vol.~41, no.~11, pp. 2568--2583, 2018.

\bibitem{weng2020early}
J.~Weng, X.~Jiang, W.-L. Zheng, and J.~Yuan, ``Early action recognition with
  category exclusion using policy-based reinforcement learning,'' \emph{IEEE
  Transactions on Circuits and Systems for Video Technology}, vol.~30, no.~12,
  pp. 4626--4638, 2020.

\bibitem{chen2018neural}
R.~T. Chen, Y.~Rubanova, J.~Bettencourt, and D.~K. Duvenaud, ``Neural ordinary
  differential equations,'' \emph{Advances in neural information processing
  systems}, vol.~31, 2018.

\bibitem{liu2019ntu}
J.~Liu, A.~Shahroudy, M.~Perez, G.~Wang, L.-Y. Duan, and A.~C. Kot, ``Ntu rgb+
  d 120: A large-scale benchmark for 3d human activity understanding,''
  \emph{IEEE transactions on pattern analysis and machine intelligence},
  vol.~42, no.~10, pp. 2684--2701, 2019.

\bibitem{wang2014cross}
J.~Wang, X.~Nie, Y.~Xia, Y.~Wu, and S.-C. Zhu, ``Cross-view action modeling,
  learning and recognition,'' in \emph{Proceedings of the IEEE conference on
  computer vision and pattern recognition}, 2014, pp. 2649--2656.

\bibitem{kipf2016semi}
T.~N. Kipf and M.~Welling, ``Semi-supervised classification with graph
  convolutional networks,'' \emph{arXiv preprint arXiv:1609.02907}, 2016.

\bibitem{li2018spatio}
C.~Li, Z.~Cui, W.~Zheng, C.~Xu, and J.~Yang, ``Spatio-temporal graph
  convolution for skeleton based action recognition,'' in \emph{Proceedings of
  the AAAI conference on artificial intelligence}, vol.~32, no.~1, 2018.

\bibitem{liu2018online}
B.~Liu, Z.~Ju, N.~Kubota, and H.~Liu, ``Online action recognition based on
  skeleton motion distribution,'' in \emph{29th British Machine Vision
  Conference}.\hskip 1em plus 0.5em minus 0.4em\relax British Machine Vision
  Association, 2018.

\bibitem{xu2019temporal}
M.~Xu, M.~Gao, Y.-T. Chen, L.~S. Davis, and D.~J. Crandall, ``Temporal
  recurrent networks for online action detection,'' in \emph{Proceedings of the
  IEEE/CVF international conference on computer vision}, 2019, pp. 5532--5541.

\bibitem{stergiou2022temporal}
A.~Stergiou and D.~Damen, ``Temporal progressive attention for early action
  prediction,'' \emph{arXiv preprint arXiv:2204.13340}, 2022.

\bibitem{rubanova2019latent}
Y.~Rubanova, R.~T. Chen, and D.~K. Duvenaud, ``Latent ordinary differential
  equations for irregularly-sampled time series,'' \emph{Advances in neural
  information processing systems}, vol.~32, 2019.

\bibitem{park2021vid}
S.~Park, K.~Kim, J.~Lee, J.~Choo, J.~Lee, S.~Kim, and E.~Choi, ``Vid-ode:
  Continuous-time video generation with neural ordinary differential
  equation,'' in \emph{Proceedings of the AAAI Conference on Artificial
  Intelligence}, vol.~35, no.~3, 2021, pp. 2412--2422.

\bibitem{chi2023adamsformer}
H.-g. Chi, K.~Lee, N.~Agarwal, Y.~Xu, K.~Ramani, and C.~Choi, ``Adamsformer for
  spatial action localization in the future,'' in \emph{Proceedings of the
  IEEE/CVF Conference on Computer Vision and Pattern Recognition}, 2023, pp.
  17\,885--17\,895.

\bibitem{agarap2018deep}
A.~F. Agarap, ``Deep learning using rectified linear units (relu),''
  \emph{arXiv preprint arXiv:1803.08375}, 2018.

\bibitem{vaswani2017attention}
A.~Vaswani, N.~Shazeer, N.~Parmar, J.~Uszkoreit, L.~Jones, A.~N. Gomez,
  {\L}.~Kaiser, and I.~Polosukhin, ``Attention is all you need,''
  \emph{Advances in neural information processing systems}, vol.~30, 2017.

\bibitem{ba2016layer}
J.~L. Ba, J.~R. Kiros, and G.~E. Hinton, ``Layer normalization,'' \emph{arXiv
  preprint arXiv:1607.06450}, 2016.

\bibitem{girdhar2021anticipative}
R.~Girdhar and K.~Grauman, ``Anticipative video transformer,'' in
  \emph{Proceedings of the IEEE/CVF International Conference on Computer
  Vision}, 2021, pp. 13\,505--13\,515.

\bibitem{han2019video}
T.~Han, W.~Xie, and A.~Zisserman, ``Video representation learning by dense
  predictive coding,'' in \emph{Proceedings of the IEEE/CVF International
  Conference on Computer Vision Workshops}, 2019, pp. 0--0.

\bibitem{han2020memory}
------, ``Memory-augmented dense predictive coding for video representation
  learning,'' in \emph{European conference on computer vision}.\hskip 1em plus
  0.5em minus 0.4em\relax Springer, 2020, pp. 312--329.

\bibitem{szegedy2016rethinking}
C.~Szegedy, V.~Vanhoucke, S.~Ioffe, J.~Shlens, and Z.~Wojna, ``Rethinking the
  inception architecture for computer vision,'' in \emph{Proceedings of the
  IEEE conference on computer vision and pattern recognition}, 2016, pp.
  2818--2826.

\bibitem{paszke2017automatic}
A.~Paszke, S.~Gross, S.~Chintala, G.~Chanan, E.~Yang, Z.~DeVito, Z.~Lin,
  A.~Desmaison, L.~Antiga, and A.~Lerer, ``Automatic differentiation in
  pytorch.''

\bibitem{mao2019learning}
W.~Mao, M.~Liu, M.~Salzmann, and H.~Li, ``Learning trajectory dependencies for
  human motion prediction,'' in \emph{Proceedings of the IEEE/CVF International
  Conference on Computer Vision}, 2019, pp. 9489--9497.

\bibitem{pang2019dbdnet}
G.~Pang, X.~Wang, J.-F. Hu, Q.~Zhang, and W.-S. Zheng, ``Dbdnet: learning
  bi-directional dynamics for early action prediction,'' in \emph{Proceedings
  of the 28th International Joint Conference on Artificial Intelligence}, 2019,
  pp. 897--903.

\bibitem{shi2015convolutional}
X.~Shi, Z.~Chen, H.~Wang, D.-Y. Yeung, W.-K. Wong, and W.-c. Woo,
  ``Convolutional lstm network: A machine learning approach for precipitation
  nowcasting,'' \emph{Advances in neural information processing systems},
  vol.~28, 2015.

\end{thebibliography}
